\documentclass{article}

\PassOptionsToPackage{numbers, compress}{natbib}

\usepackage[eandd,preprint]{neurips_2026}

\usepackage[utf8]{inputenc} %
\usepackage[T1]{fontenc}    %
\usepackage{hyperref}       %
\usepackage{url}            %
\usepackage{booktabs}       %
\usepackage{amsfonts}       %
\usepackage{nicefrac}       %
\usepackage{microtype}      %
\usepackage{xcolor}         %
\usepackage{xspace}
\usepackage{listings}
\usepackage[most]{tcolorbox}
\usepackage{enumitem}
\usepackage[table]{xcolor}
\usepackage{tabularx}
\usepackage{makecell}
\usepackage[table]{xcolor}  %
\usepackage{threeparttable}

\usepackage{titlesec}
\usepackage{microtype}
\titlespacing*{\section}{0pt}{0.8ex}{0.4ex}
\titlespacing*{\subsection}{0pt}{0.6ex}{0.3ex}
\titlespacing*{\subsubsection}{0pt}{0.4ex}{0.2ex}
\titlespacing*{\paragraph}{0pt}{0.5ex}{1ex}

\usepackage[skip=4pt]{caption}

\setlist{nosep}                  %
\renewcommand{\arraystretch}{0.95}

\definecolor{tdcwrong}{HTML}{C53030}
\definecolor{litright}{HTML}{276749}
\definecolor{stccolor}{HTML}{B8860B}

\newcommand{\pmids}[1]{\hspace{0.4em}{\scriptsize\color{black!55}PMID #1}}
\usepackage[capitalise]{cleveref}

\newcommand{\ourmethod}{\texttt{Starling}\xspace}

\newcommand{\totalextract}{$\sim6.3$M\xspace}

\definecolor{jsonbg}{HTML}{F7F7F8}
\definecolor{jsonframe}{HTML}{D0D0D6}
\definecolor{jsonkey}{HTML}{2E5CB8}
\definecolor{jsonstring}{HTML}{B5500B}
\definecolor{jsoncomment}{HTML}{8B8B99}
\definecolor{jsonnull}{HTML}{7A3E9D}

\lstdefinelanguage{jsonschema}{
  basicstyle=\ttfamily\small,
  columns=fullflexible,
  keepspaces=true,
  showstringspaces=false,
  escapeinside={(*}{*)},
  literate=
    *{null}{{{\color{jsonnull}\textbf{null}}}}{4}
     {[}{{{\color{black}[}}}{1}
     {]}{{{\color{black}]}}}{1}
     {\{}{{{\color{black}\{}}}{1}
     {\}}{{{\color{black}\}}}}{1},
  morestring=[b]",
  stringstyle=\color{jsonstring},
  comment=[l]{\#},
  commentstyle=\color{jsoncomment}\itshape,
}

\newtcblisting{jsonfig}[1][]{%
  listing only,
  listing options={language=jsonschema},
  colback=jsonbg,
  colframe=jsonframe,
  boxrule=0.5pt,
  arc=2pt,
  left=4pt, right=4pt, top=0pt, bottom=0pt,
  boxsep=0pt
  #1
}

\title{Self Driving Datasets: From 20 Million Papers to Nuanced Biomedical Knowledge at Scale}

\author{%
  Haydn Jones\textsuperscript{1} \quad
  Yimeng Zeng\textsuperscript{1} \quad
  Alden Rose\textsuperscript{1} \quad
  Li S. Yifei\textsuperscript{1} \quad
  Yining Huang\textsuperscript{1} \\[2pt]
  Kaiwen Wu\textsuperscript{1} \quad
  Jiaming Liang\textsuperscript{1} \quad
  Maggie Ziyu Huan\textsuperscript{1} \quad
  Yoseph Barash\textsuperscript{2} \quad
  Cesar de la Fuente-Nunez\textsuperscript{3} \\[2pt]
  Osbert Bastani\textsuperscript{1} \quad
  Zachary Ives\textsuperscript{1} \quad
  Mark Yatskar\textsuperscript{1,$\ast$} \quad
  Jacob R. Gardner\textsuperscript{1,$\ast$} \\[6pt]
  \normalfont\textsuperscript{1}Department of Computer and Information Science, University of Pennsylvania \\
  \normalfont\textsuperscript{2}Department of Genetics, University of Pennsylvania \\
  \normalfont\textsuperscript{3}Departments of Bioengineering and Chemical and Biomolecular Engineering, \\
  \normalfont University of Pennsylvania \\[4pt]
  \normalfont\textsuperscript{$\ast$}Equal contribution
}

\begin{document}

\maketitle

\begin{abstract}
Manually curated biomedical repositories---spanning bioactivity, genomics, and chemistry---are expensive to maintain, lag behind primary literature, and often discard experimental context. The absence of contextual information obscures critical nuances, thereby complicating the assessment of data correctness and coverage, necessary criteria for building high-quality models. We show that PubMed itself can be turned into structured datasets---autonomously and cost-effectively---that are larger, more nuanced, and more accurate than the curated databases they would replace. We present three coupled contributions: (1) an LLM-based entity-tagging pipeline, grounded in nine biomedical ontologies, that tags 4.5 billion entities across 19 categories in a 22.5M-paper, 2.5-trillion-token PubMed corpus; (2) hybrid sparse–dense retrieval infrastructure supporting surgical entity-filtered semantic queries over the tagged corpus; and (3) \ourmethod{}, a multi-agent deep research system that, given only a natural-language task description, autonomously designs precision- and recall-targeted retrieval filters, induces an extraction schema, and emits structured records with nuance-rich fields and supporting passages. Applied to six tasks---blood-brain barrier permeability, oral bioavailability, acute toxicity (LD50), gene-disease associations, protein subcellular localization, and chemical reactions---\ourmethod{} produces \totalextract records (per-task scale ranges from 91K to 3M); several of these are, to our knowledge, the largest public datasets for their respective properties. Frontier-model rejection of our kept extractions is 0.6–7.7\% across tasks, surprisingly far below the error rates we measure on the widely used, manually curated counterparts (e.g., 16.5\% on \texttt{BBB\_Martins}, 7.3\% on \texttt{Bioavailability\_Ma}). Beyond scale and accuracy, the attached supporting passages carry nuance that tabular databases discard: for example, oral bioavailability of a molecule might depend on whether the patient is fed or fasted. Together, the corpus, retrieval layer, and agent establish a foundation for multimodal predictive and generative models in AI-driven therapeutic design. We release code and links to our datasets at \url{https://anonymous.4open.science/r/starling-B026/}.
\end{abstract}

\section{Introduction}
\label{sec:intro}

For decades, the biomedical sciences have invested heavily in the collection, curation, and dissemination of experimental data. This investment has paid off: centralized repositories have enabled rapid and widely successful applications of AI and machine learning across genomics, protein folding, medical and laboratory imaging, and drug discovery~\cite{alphafold,Johnson2023MIMICIVAF,Wan2024MachineLF}. However, much of this data is captured manually at tremendous cost and aggregated into structured, tabular databases~\cite{huang2022tdc,kim_pubchem_2016,kearnes_open_2021}. The dominant use case for these collections is the familiar pattern of supervised learning: given input features, predict a label. Even recent generative applications are primarily unimodal: given a protein's primary structure, generate its folded structure or a novel binder.

The success of this paradigm is difficult to overstate. But the ability of modern large language models to extract nuanced information directly from unstructured text~\cite{obeidat_llms_2025,majid_evaluating_2025,zhou_universalner_2024} exposes three shortcomings of existing biomedical data repositories that are increasingly difficult to ignore:
\begin{enumerate}[leftmargin=*, itemsep=0pt, topsep=0pt]
    \item \emph{Existing data lacks nuance.} Most medicinally relevant properties are not functions of the molecule alone. Oral bioavailability depends on whether a drug is taken on an empty stomach, the formulation it is delivered in, and the population it is dosed in; reaction yields depend on procedure and conditions; assay readouts depend on cell line and protocol. Models that operate unimodally on the molecule can at best estimate the marginal distribution over these conditions. Text makes the conditions themselves first-class.

    \item \emph{Existing data is incomplete.} Repositories lag the literature by years, and even at steady state they capture a small fraction of what has been measured: datasets could plausibly grow by orders of magnitude. Worse, evidence for the same physical phenomenon is often reported in incompatible forms. Hepatotoxicity may surface as binding to a liver enzyme or as a population-level adverse-event signal; oral bioavailability may be reported in absolute terms or relative to a reference compound. These are awkward to flatten into a single schema but easy to describe in language.

    \item \emph{Existing data is less accurate than commonly assumed.} When we audit widely used benchmarks against human judgment (\cref{sec:results-accuracy}), we find non-trivial label error: $16.5\%$ on TDC BBB and $7.3\%$ on TDC Oral Bioavailability~\cite{huang2022tdc}. Models trained on these labels inherit that error as a noise floor and a ceiling on achievable test performance.
\end{enumerate}
The natural response to these shortcomings is to go back to the source. The primary literature contains, in principle, both the missing data and the experimental context that tabular schemas discard---and unlike a curated database, it grows continuously. Historically, the obstacle has been access: extracting structured records from tens of millions of papers requires a kind of surgical retrieval that neither classical information extraction nor recent deep research agents are equipped to perform. General-purpose deep research agents are designed to synthesize relevant documents into a narrative report, not to exhaustively sweep a corpus for every instance of a phenomenon and emit a structured record~\cite{huang2025biomni,Mitchener2025KosmosAA}.

In this paper we present a system that does exactly this. It operates over a tagged and indexed corpus of $22.5$ million PubMed papers and answers natural-language extraction tasks of the form \emph{``find all examples in the corpus where the oral bioavailability of a small molecule is discussed''} or \emph{``find all instances where evidence of any gene-disease association is presented.''} Given only a task description, the system autonomously designs retrieval filters combining entity constraints and semantic queries, induces an extraction schema, and outputs structured records linked to supporting passages in the source paper. This enables the construction of large, nuanced, literature-grounded datasets in days, replacing years of manual curation, and our audits show this can be done at even higher accuracy.

Our specific contributions are: 
\begin{enumerate}[leftmargin=*, itemsep=2pt, topsep=2pt]
    \item \textbf{A method for self-driving biomedical dataset construction.} We introduce \texttt{Starling}, a multi-agent deep research system that autonomously designs and extracts structured records from the literature. \texttt{Starling} operates over a tagged and indexed corpus, supported by two components we develop alongside it: an entity tagging model fine-tuned from \texttt{gpt-oss-20b}~\cite{gptoss} covering $19$ top-level biomedical categories, grounded in nine reference ontologies and, for small molecules and proteins, directly to SMILES and amino acid sequences; and a hybrid sparse--dense retrieval layer that exposes the tagged corpus through entity-filtered semantic queries.

    \item \textbf{Evidence that automated extraction can exceed manual curation in size, accuracy, and nuance.} Across six therapeutic discovery tasks, \texttt{Starling} yields \totalextract judge-filtered records, including several that are, to our knowledge, the largest public datasets for their properties. Frontier-model and human audits estimate per-task error at $0.6$--$4.7\%$, below our estimates for widely used manually curated benchmarks. Supporting passages add signal beyond structure-only representations: for a given molecule, residual variation in oral bioavailability is often significantly explained by experimental context (e.g., fed vs. fasted dosing). Existing deep research systems either fail on these extraction tasks or mainly rediscover scientist-curated datasets.

    \item \textbf{Open release of the system, recipes, and datasets.} We open-source the \ourmethod agent, entity tagging model, retrieval infrastructure, and end-to-end recipes for reproducing the pipeline on any suitably licensed corpus, and provide open access to the system running on our own licensed corpus. We also release six datasets covering small-molecule blood-brain barrier permeability, oral bioavailability, acute toxicity (LD$50$), gene-disease associations, protein subcellular localization, and chemical reactions. See \cref{sec:open_access_plan} for details on our release plans.
\end{enumerate}

\begin{figure}
    \centering
    \includegraphics[width=\linewidth]{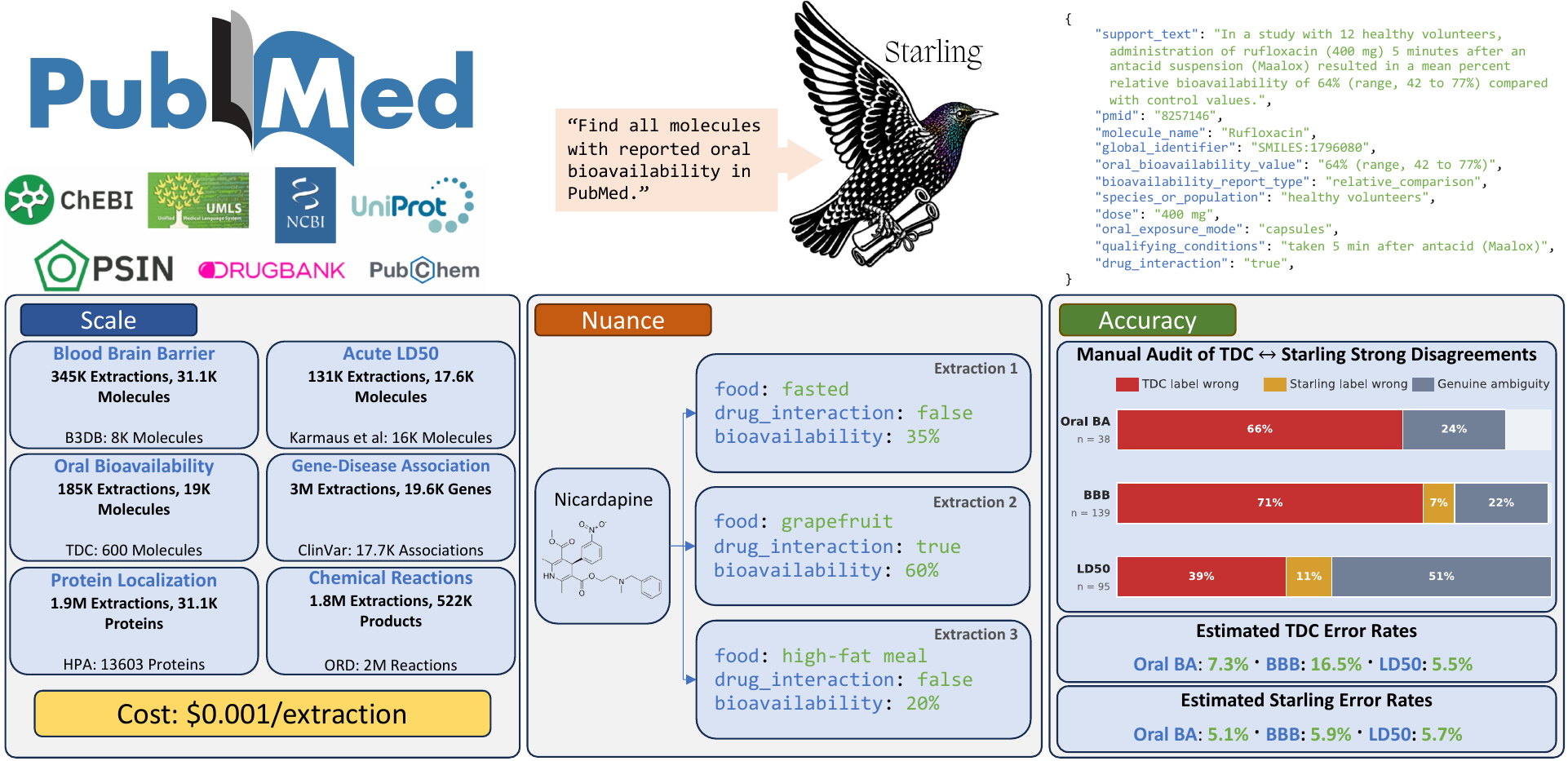}
    \caption{\ourmethod is a promptable agentic system that automatically creates datasets from biomedical literature. 
\ourmethod is fully self-driving, requiring no human intervention, and produces datasets at comparable or larger scale than manual resources, capturing nuanced context often missing in existing datasets, with similar or better accuracy.
\ourmethod can keep datasets better synchronized to quickly expanding information in literature, at under 1 cent per datum it extracts. 
    }
    \label{fig:starling_overview}
\end{figure}

\section{Corpus and Database Construction}
\label{sec:corpus}
We use a corpus of 22.5 million PubMed articles. This corpus represents the subset of PubMed available to us with (1) journal licensing terms that do not preclude AI usage and (2) relatively straightforward pathways to acquire PDFs. We extract text and tables into markdown files using \texttt{dots.ocr} \citep{li2025dotsocrmultilingualdocumentlayout}. In this work, we do not yet consider the information contained in figures. This results in a corpus of approximately 2.5 trillion tokens, which we divide into overlapping chunks of 5 paragraphs and embed with \texttt{Qwen3-Embedding-4B} \citep{zhang_qwen3_2025}, creating an initial database of roughly 250 million 1024-dimensional embedding vectors. In contrast to traditional RAG and deep research agents \citep{lala_paperqa_2023, team_tongyi_2025,skarlinski_language_2024}, our system uses these embeddings to \textit{re-rank} search results rather than filter the corpus.

\paragraph{Challenges for self-driving datasets.} Consider an extraction task such as ``find a large set of molecules for which blood-brain barrier permeability is discussed.'' At 2.5 trillion tokens, a linear pass over the full corpus per task is prohibitively expensive, even with a fast LLM, ruling out brute-force extraction. Existing deep research agents \citep{lala_paperqa_2023, team_tongyi_2025,skarlinski_language_2024} synthesize reports from a small set of retrieved documents rather than exhaustively scanning a corpus. Embedding-based retrieval can surface relevant passages, but at corpus scale it is a noisy \emph{filter} and works better as a passage \emph{re-ranker}. We need a cheap filtering mechanism that narrows tens of millions of papers to manageable subsets likely to contain the target data.

Entity tagging enables this: for molecule-related tasks, we focus only on chunks that mention at least one small molecule, discarding most of the corpus. However, extraction tasks are not known in advance—a future query may seek gene-disease associations or reaction yields, each needing different entity co-occurrences. The tagging layer must therefore span a broad range of biomedical entity types so downstream agents can compose arbitrary entity filters on demand.

\subsection{Entity Tagging}
\label{sec:ner}

We tag the corpus with 19 top-level biomedical entity types---\texttt{SmallMolecule}, \texttt{Gene}, \texttt{Disease}, \texttt{Protein}, and others (see \cref{app:corpus-details})---grounded in nine reference ontologies. Recent work shows that LLMs can outperform BERT-style taggers on biomedical NER \citep{obeidat_llms_2025,majid_evaluating_2025,zhou_universalner_2024}, motivating the teacher–student pipeline below. Each paragraph in the resulting corpus carries canonical, ontology-linked entity tags that downstream agents can compose into arbitrary entity filters.

\paragraph{Tag schema.} The teacher labels a window of 3 consecutive paragraphs with a list of entity tags. Each tag includes a disambiguated canonical name selected by the model, the entity type, optional synonyms to aid normalization, the exact surface forms in the text, and the subset of paragraphs in which the entity is mentioned. Type-specific optional fields capture organism specificity (e.g., \textit{Homo sapiens} for genes and proteins), medicinal-chemistry aliases (e.g., \texttt{6f}), and resolved names (e.g., the IUPAC name for a small molecule).

\paragraph{Normalization.} After tagging, each extracted entity is grounded to a reference ontology. UMLS \citep{bodenreider_unified_2004} is our default, covering \texttt{Disease}, \texttt{Anatomy}, \texttt{Phenotype}, \texttt{GOTerm}, and most other concept types via aggregated biomedical ontologies; when using other ontologies, we follow the UMLS free-text–to–concept matching pipeline\footnote{See \url{https://www.nlm.nih.gov/research/umls/implementation_resources/query_diagrams/index.html} for details on UMLS term normalization.}. Where more authoritative domain resources exist, we instead use a cascade of OPSIN \citep{lowe_chemical_2011}, DrugBank \citep{knox_drugbank_2024}, ChEBI \citep{degtyarenko_chebi_2008}, and PubChem \citep{kim_pubchem_2016} (in that order) for \texttt{SmallMolecule}; ChEBI for \texttt{SmallMoleculeClass}; NCBI Taxonomy \citep{schoch_ncbi_2020} for \texttt{Organism}; UniProt \citep{apweiler_uniprot_2004} for \texttt{Protein}; and NCBI Gene \citep{brown_gene_2015} for \texttt{Gene}.

\paragraph{Distillation pipeline.} We sample 50{,}000 papers and run gpt-oss-120b \citep{gptoss} 19 times per window—once per entity type—using a standardized, type-specific prompt with its own rules and few-shot examples. Running separate single-type passes lets the model focus on one category at a time, improving precision and recall over a joint 19-type pass. Before training the student model, we validate the 1{,}000 most frequent entities per type by mapping them to a reference ontology and selecting the 10 most likely candidates. GPT-5 \citep{singh_openai_2025} then chooses the correct candidate (or rejects all), and we replace the teacher’s tags with the ontology’s preferred names and synonyms, enforcing consistent nomenclature for high-frequency concepts. We then merge tags across types within each window to obtain windows labeled with all 19 types and fine-tune \texttt{gpt-oss-20b} on this combined supervision. The distilled student is run on the full 22.5M-paper corpus and its outputs are re-normalized (without GPT-5 validation, which is used only at distillation time), as the larger corpus surfaces entities missing from the 50{,}000-paper sample.

\subsection{Constructing Entity Filters} 
\label{sec:entity-filters}

The deep research agents we construct next interact with our tagged corpus primarily by using \textit{entity filters} that select subsets of the corpus, which we then re-rank with a semantic query---for example, ``search only over windowed chunks in the corpus that mention both some small molecule and the blood-brain barrier.'' Each filter is represented as a \texttt{FilterSpec} containing a CNF entity constraint (outer conjunction, inner disjunction, with negation support via prefix notation). We provide a simple example of a filter specification and a discussion of how it is used in \cref{app:corpus-details}.

\section{Deep Research Agents for Self-driving Datasets}
\label{sec:agent}

To convert the tagged corpus into structured datasets, we introduce \ourmethod, a multi-agent deep research system that turns a natural-language task description into a structured dataset, requiring no user-specified schema and no manual paper selection. 

Literature-scale extraction faces two main challenges. First, we cannot make linear passes over the 2.5T-token corpus for each task, so we must choose which subset of the literature to process and in what order. Second, if the LLM is free to invent fields on each extraction, hundreds of thousands of extractions will produce an unusably inconsistent schema. Thus, beyond universal fields like \texttt{PMID} and \texttt{support\_text}, we must predefine a set of relevant structured fields (e.g., \texttt{fed\_or\_fasted} for oral bioavailability) based on domain knowledge. To address this, \ourmethod{} runs in two phases. During \emph{query construction}, we filter and reorder the corpus and inspect matching papers to design the extraction schema. During \emph{extraction}, we process this subcorpus to perform the extractions.

\subsection{Phase 1: Query Construction}
\label{sec:corpus-filter}

A \texttt{Proposer} agent constructs a set of \textbf{probes} $\{P_1,...,P_T\}$. Each probe consists of (1) an entity filter (\cref{sec:entity-filters}) and (2) a semantic query. Each probe ideally targets a distinct yet relevant portion of the corpus, achieving high precision and recall on minimally overlapping subsets. The agent explores the corpus with tools that run semantic search on entity-filtered subsets to form initial probe specifications, then iteratively creates probes to maximize estimated precision and recall.

\paragraph{Maximizing precision.} Each text matched by probe $P_{i}$ should yield a relevant extraction as often as possible. To estimate this precision, the agent samples papers from a probe's filter and tasks a \texttt{Validator} agent to judge what fraction is relevant. An \texttt{Investigator} agent examines text that the \texttt{Validator} deems irrelevant but passes the filter. It suggests strategies to refine the probe that would exclude these irrelevant results while minimally impacting relevant ones.

\paragraph{Maximizing recall.} The union of our probes should ideally filter out minimally relevant text. To estimate recall, we need to check how often we can find relevant extractions in papers that our probes currently \textit{exclude}. To do this, we look at chunks that are highly relevant according to the $T$ semantic queries but are excluded by all $T$ entity filters. The \texttt{Validator} judges each chunk, and the fraction marked relevant is our \textbf{recall-gap estimate}. If this fraction is high, it means it is ``easy'' to find relevant text that our current probe set excludes. To fix this, the \texttt{Proposer} must either make some probe more permissive or construct a new probe $P_{t+1}$, guided by \texttt{Investigator} suggestions, that captures relevant text missed by $P_{1},...,P_{t}$.

This process repeats. At each iteration, the \texttt{Proposer} uses (1) precision and recall estimates, and (2) \texttt{Investigator} feedback to modify existing probes or construct additional probes to capture the gaps identified. We continue refining until we estimate 80\% precision and at most a 15\% recall gap, or until a maximum number of probes is reached. See \cref{app:agent-details} for more details.

\paragraph{Schema induction.} After selecting probes $P_{1},...,P_{T}$, we define a shared extraction schema for all dataset rows. The \texttt{Proposer} samples a few retrieval-set papers and runs the \texttt{Extractor} (described below) \emph{without} a fixed schema. From the resulting free-form records, the agent derives a unified schema with field names, types, allowed values, and a concrete \emph{task instantiation} specifying what counts as a valid record, what to ignore, and what default assumptions to use. For the blood-brain barrier task, the instantiation requires a named compound and an explicit permeability statement, assuming an intact barrier by default (so a diseased state becomes a qualifying condition).
The \texttt{Validator} scores extraction quality under the proposed schema; if it is below target, the \texttt{Proposer} revises and the loop repeats. Once quality meets the target, the schema and instantiation are frozen and applied to the full retrieval set. 
At this stage, a human can intervene to shape the probes or extraction schema (e.g., in our \texttt{Reactions} task, we skip schema construction and use a hand-authored schema that roughly matches the one used by the Open Reaction Database in \cref{sec:results} ).

\subsection{Phase 2: Extraction}
\label{sec:extraction}

Once the \texttt{Proposer} returns a retrieval specification, we execute it against the corpus to produce the set of matching papers and then rank papers by (1) semantic similarity and (2) the number of probe filters they match. Extraction is substantially more straightforward than planning: an \texttt{Extractor} agent loops over windows from the ordered subcorpus implied by the probe set and emits extractions using the instructions and schema that were generated during query construction.

Because the \texttt{Extractor} runs once per window without a global view, the deduplicated set still includes off-task, mis-attributed, or low-fidelity rows. A \texttt{Judge} filters these as the final pipeline stage. After completing the extraction pipeline for a task, we use GPT-5.4 to score a small sample of extractions along five axes: \emph{task relevance} (the record matches the task), \emph{primary-label correctness} (the headline value matches the source), \emph{span fidelity} (the supporting passage contains the claim), \emph{secondary-field accuracy} (auxiliary fields match the passage), and \emph{entity attribution} (the primary entity is correctly identified and resolved). We then distill these judgments into a Qwen3.5-9B model using SFT and GRPO \citep{shao_deepseekmath_2024} across all six tasks. A record is kept only if it passes all five axes. \cref{sec:results-accuracy} reports per-task pass rates and the rate at which frontier models reject kept entries. See \cref{app:agent-details} for more details on the judgment pipeline.

\section{Experimental Setup}

\paragraph{Case Studies.} We evaluate \ourmethod{} on six therapeutically relevant tasks (\cref{fig:task-prompts}): three small-molecule properties---blood-brain barrier permeability (\texttt{BBB}), oral bioavailability (\texttt{Oral}), and \texttt{LD50}---plus gene-disease associations (\texttt{GDA}), protein subcellular localization (\texttt{PSL}), and chemical reactions (\texttt{Reactions}). Each dataset record includes a primary entity, a target measurement, structured task-specific fields, and a free-text supporting passage. While the \texttt{Extractor} (\cref{sec:extraction}) induces schemas autonomously, for \texttt{Reactions} we manually authored the schema and per-field extraction guidance to enable direct comparisons with ORD. 

\paragraph{Baselines.}
To assess the reliability of \ourmethod{} datasets, we align each case study with an existing benchmark: TDC for BBB, Oral, and LD50~\citep{huang2022tdc}; B3DB for BBB~\citep{meng_curated_2021}; ClinVar for GDA~\citep{landrum2018clinvar}; UniProt (\textit{Homo sapiens} only) for PSL~\citep{apweiler_uniprot_2004}; and the Open Reaction Database (ORD) for Reactions~\citep{kearnes_open_2021}. We compare these on accuracy, size, and cost, and benchmark against existing deep research systems in \cref{sec:results-novelty} and \cref{app:deep_research_agents}. We also quantify the downstream effects of training with \ourmethod{} data. Our base predictor is MiniMol~\citep{klaser_textttminimol_2024}, a supervised model trained on TDC data and outputs from a physical simulator that is currently state-of-the-art on a range of tasks.

\section{Results}
\label{sec:results}

We organize the rest of this section around three main claims: \ourmethod's extractions are large in scale and effective as training data (\cref{sec:results-size}), accurate---and even \emph{more accurate} than the manually curated benchmarks they extend (\cref{sec:results-accuracy}), and outperform existing deep-research agents for dataset curation while being cost-effective (\cref{sec:results-novelty}).

\subsection{\ourmethod Datasets are Large-Scale and Drive Effective Model Training}
\label{sec:results-size}
\begin{table}[!ht]
    \centering
    \caption{Per-task extraction counts collected by \ourmethod{}, alongside error rates from two strong LLM evaluators measuring the fraction of extractions that fail at least one evaluation axis.}
    \label{tab:datasets}
    \vspace{1em}
    \rowcolors{2}{white}{gray!6}
    \small
    \begin{tabular}{lrrrrrrr}
        \toprule
         & \multicolumn{3}{c}{\ourmethod{} Statistics} & \multicolumn{2}{c}{Dataset Comparisons} & \multicolumn{2}{c}{Error Rate} \\
        \cmidrule(lr){2-4}
        \cmidrule(lr){5-6}
        \cmidrule(lr){7-8}
        Task      & Extractions & Papers & Unique IDs & Coverage          & Scale       & Opus 4.7 & GPT-5.4 \\
        \midrule
        BBB       &  304K       &   122K &  31.1K     &    44.3\% TDC     & 4x B3DB     &   4.05\% &  5.88\% \\
        GDA       & 3.00M       &   145K &  19.6K     &    41.9\% ClinVar & 6x ClinVar  &   0.88\% &  5.70\% \\
        LD50      &   91K       &  30.3K &  17.6K     &    31.5\% TDC     & 2x TDC      &   3.89\% &  5.72\% \\
        Oral      &  163K       &  45.0K &  19.0K     &    96.1\% TDC     & 29x TDC     &   0.56\% &  5.06\% \\
        PSL       & 1.79M       &   171K &  36.1K     &    90.8\% UniProt & 24x UniProt &   1.12\% &  5.56\% \\
        Reactions &  916K       &   259K &   545K     &      ---          & 0.51x ORD   &   4.68\% &  7.78\% \\
        \bottomrule
    \end{tabular}
\end{table}

\paragraph{\ourmethod{} Dataset Sizes.} In \cref{tab:datasets} we report, for each case study, the total number of extractions \ourmethod produces and how they compare to existing resources. We compare datasets in two ways: (1) how many instances from existing resources we cover (Coverage), and (2) overall size (Scale). Across all case studies, \ourmethod produces datasets as large as existing resources while maintaining high coverage. While the literature does not cover every instance in the manually curated resources, \ourmethod's extractions remain more up to date than these resources.
In reactions, we discover nearly half a million new products not yet described in public datasets (\cref{sec:reactions:results}).

\begin{table}[t]
    \centering
    \caption{Predictive performance on (1) the TDC test set with and without additional training data extracted from the literature, and (2) on a new literature extracted test set}
    \label{tab:admet-results}
    \vspace{1em}
    \small
  \rowcolors{2}{white}{gray!6}
    \setlength{\tabcolsep}{4pt}
    \begin{tabular}{@{}l l | c c | c c@{}}
        \toprule
        \textbf{Task} & \textbf{Metric} & \textbf{TDC Test} & \textbf{TDC Test (Aug.)} & \textbf{Lit. Test} & \textbf{Lit. Test (Aug.)} \\
        \midrule
        Oral Bioavailability                                        & AUROC $\uparrow$ & 0.692 & \textbf{0.779} & 0.768 & \textbf{0.821} \\
        LD50                                                         & MAE $\downarrow$ & 0.602 & \textbf{0.575} & 0.650 & \textbf{0.641} \\
        BBB                                                          & AUROC $\uparrow$ & \textbf{0.916} & 0.909 & 0.768 & \textbf{0.890} \\
        \bottomrule
    \end{tabular}
\end{table}

\paragraph{Using \ourmethod{} datasets to augment property prediction.} In \cref{tab:admet-results}, we consider how useful \ourmethod-curated data is for training and evaluating models. Starting with MiniMol~\citep{klaser_textttminimol_2024}, we first fine-tune it on manually curated TDC datasets and evaluate on TDC test data or \ourmethod-derived test data. Then, we evaluate the effect of augmenting training data with \ourmethod instances chosen using a train/test splitting mechanism similar to TDC's (Aug), comparing test performance with and without augmentation. Across the three TDC-backed case studies, augmenting predictor training with \ourmethod data never substantially hurts performance and often improves it, both on diverse literature-sourced data and on manually curated test sets.

\paragraph{Cost.} Across the six tasks, end-to-end \ourmethod extractions cost approximately $\$0.001$ per record on average, dominated by inference costs for extraction. Distilling and training the judge was a one-time cost of approximately \$400, amortized across the six tasks.

\subsection{\ourmethod{} Datasets Are Accurate and Often Surpass Manual Curation}
\label{sec:results-accuracy}
\begin{table}[ht]
  \centering
  \small
  \caption{Human-validated label error rates on molecules in four manually curated databases where \ourmethod{} found $\geq 3$ extractions. We hand-judged the disagreement subset (majority of \ourmethod{} extractions disagree with the database label, or $> 0.5$ log units of disagreement for LD50), reviewing 8{,}765 extractions in total. The \textbf{confirmed-wrong} columns considers all $\geq 3$-extraction molecules; e.g., the B3DB row reflects 1086 total molecules, of which $214/1086 = 19.7\%$ were judged mislabeled. Both rates are lower bounds: molecules whose \ourmethod{} extractions agreed with the database label were not audited and are presumed correct.}
  \vspace{1em}
  \label{tab:human-validation}
  \setlength{\tabcolsep}{4pt}
  \rowcolors{2}{white}{gray!6}
  \begin{tabular}{l r r r r r r}
    \toprule
    \makecell[l]{Baseline\\Dataset} & \makecell[r]{Mols.\\Reviewed}
      & \makecell[r]{Baseline\\wrong} & \makecell[r]{\ourmethod{}\\wrong}
      & \makecell[r]{Lit.\\ambiguous} & \makecell[r]{Confirmed-wrong\\baseline rate} 
      & \makecell[r]{Confirmed-wrong\\\ourmethod{} rate} \\
    \midrule
    Bioavailability\_Ma & 34  & 25  & 0  & 9  & \textbf{7.3\%}  & \textbf{0.0\%} \\
    BBB\_Martins        & 138 & 98  & 10 & 30 & \textbf{16.5\%} & \textbf{1.7\%} \\
    LD50\_Zhu           & 95  & 37  & 10 & 48 & \textbf{5.5\%}  & \textbf{1.5\%} \\
    B3DB                & 316 & 214 & 20 & 78 & \textbf{19.7\%} & \textbf{1.8\%} \\
    \bottomrule
  \end{tabular}

\end{table}

\paragraph{Human calibration of the LLM judge.} We first verify that an LLM judge is a reasonable proxy for human review on the five axes (\cref{sec:extraction}). Six annotators independently reviewed \texttt{BBB} and \texttt{LD50} extractions, scoring each. For each task, 20 extractions were shared across all reviewers (for inter-annotator agreement), and each reviewer labeled 5 additional unique extractions, yielding 50 total extractions per task. Human pass rates were $97.3\%$ on \texttt{BBB} and $94.0\%$ on \texttt{LD50}, matching the frontier-judge pass rates in \cref{tab:datasets}. Judge--human agreement on the shared subset ($95.8\%$ \texttt{BBB}, $92.0\%$ \texttt{LD50}) is within about one percentage point of human--human agreement ($95.1\%$ \texttt{BBB}, $90.6\%$ \texttt{LD50}), indicating that the LLM judge is as consistent with annotators as they are with each other.

\paragraph{Judge results.} With the LLM judge thus calibrated, we evaluated at least 3{,}000 kept extractions per task with two frontier judges, reporting the rate at which each flags an extraction as failing on at least one axis (\cref{tab:datasets}). Both judges report error rates below 10\% on every task, typically at or below 5\%.

\paragraph{Comparing \ourmethod{} accuracy to existing resources.}
We compare \ourmethod{} to four manually curated databases, considering only molecules with at least three \ourmethod{} extractions. This threshold lets us label each case as correct, ambiguous, or wrong, rather than as a binary correct/incorrect. We manually reviewed molecules where the majority of \ourmethod{} extractions disagreed with the database; cases where the majority agreed were treated as confirming the database label. Annotators investigated each disagreement and read \ourmethod{}'s supporting documents to judge whether the database label was correct (\cref{tab:human-validation}). Most disagreements were database errors: every resource had a confirmed-wrong rate above 5\%, and B3DB exceeded 19\%. Using the same denominator, \ourmethod{}'s confirmed-wrong rate is $\leq 1.8\%$, indicating that its extractions are at least as accurate as these curated datasets, and often more so. For selected ``clear'' errors, see \cref{tab:greatest-hits-bbb}.

\subsection{\ourmethod{} is a Higher-Quality Data Generator Than Frontier Deep Research Systems}
\label{sec:results-novelty}
\begin{table}[h]
\centering
\caption{Literature-grounded record counts per task: the strongest deep-research baseline \emph{for that task} vs.\ \ourmethod{}. A row counts as ``literature-grounded'' if it carries a real PMID (or an upstream-dataset PMID we recovered for the agent; rescue methodology in \cref{app:main-results-notes}); we credit baselines at face value with no per-row LLM judging. \ourmethod{} counts are post-judge-filter (\cref{tab:datasets}); baseline counts are unfiltered. GPT-5.4 Deep Research is strongest on T1; Kosmos AI-Scientist is strongest on T2/T4/T5 and ties GPT-5.4 Deep Research on T3 (both consume the same PubTator3 BioREx file). See more details in \cref{app:deep_research_agents}, and full per-agent breakdown in \cref{app:results,app:main-results-notes}.}
\vspace{1em}
\label{tab:baseline-grounded-rows}
{
\small
\rowcolors{2}{white}{gray!6}
\setlength{\tabcolsep}{6pt}
\renewcommand{\arraystretch}{1.15}
\begin{tabular}{@{}llrr@{}}
\toprule
\textbf{Task} & \textbf{Strongest baseline} & \textbf{Baseline rows} & \textbf{\ourmethod{} (judge-filtered)} \\
\midrule
BBB   & GPT-5.4 Deep Research              &     10{,}990  &     304{,}845 \\
Oral  & Kosmos AI-Sci.                     &     30{,}308  &     163{,}815 \\
GDA   & Kosmos / GPT-5.4 Deep Res.$^*$     & 1{,}455{,}774 & 3{,}009{,}161 \\
LD50  & Kosmos AI-Sci.                     &     56{,}658  &      91{,}597 \\
PSL   & Kosmos AI-Sci.                     &    180{,}685  & 1{,}798{,}418 \\
\bottomrule
\end{tabular}
}
\end{table}

We evaluate how five frontier deep research systems handle dataset-construction requests: BioMNI-Phylo (a sandboxed Python coding agent)~\cite{huang2025biomni}, Claude Research Mode (Opus 4.6 Extended), GPT-5.4 Deep Research, GPT-5.4 Pro Extended Thinking, and Kosmos AI-Scientist (Edison Scientific)~\cite{Mitchener2025KosmosAA}. Many of these produce narrative construction plans rather than data, so we passed these plans to Claude Code for execution. Manual inspection reveals three main strategies: (1) reformatting existing curated databases; (2) small-scale PubMed text mining with high false-positive rates; and (3) catastrophically invalid bulk extraction at scale. None produce datasets that are simultaneously large, paper-derived, and high-precision---consistent with the fact that their agentic designs are ill-suited to this task. In \cref{tab:baseline-grounded-rows}, we compare to, for each task, the deep research system with the largest number of correctly formatted, literature-grounded instances. We assume baseline output is correct and compare on scale (\cref{app:main-results-notes,app:strategy} has per-agent strategies and statistics).

\section{\ourmethod Builds Contextualized Data Enabling Nuanced Predictions}
\label{sec:results-nuance}

\ourmethod curates rich, high-signal, property-altering context for each extraction. This context captures the conditions under which properties change, encodes human understanding of mechanisms of action, and surfaces free-text information tied to each measurement. For example, a drug's oral bioavailability can vary greatly with food intake and meal characteristics (e.g., Nicardipine in \cref{fig:starling_overview}).
Such nuance has been largely overlooked by dataset designers. 
During information extraction, \ourmethod collects two kinds of context: (1) named fields with values from a fixed set, and (2) unstructured supporting text that provides context and justifies those values.
Even in data settings that collect context, it is often very incomplete and \ourmethod systematically finds more (\cref{sec:reactions:richness}).
\begin{figure}[ht!]
    \centering
    \includegraphics[width=1\linewidth]{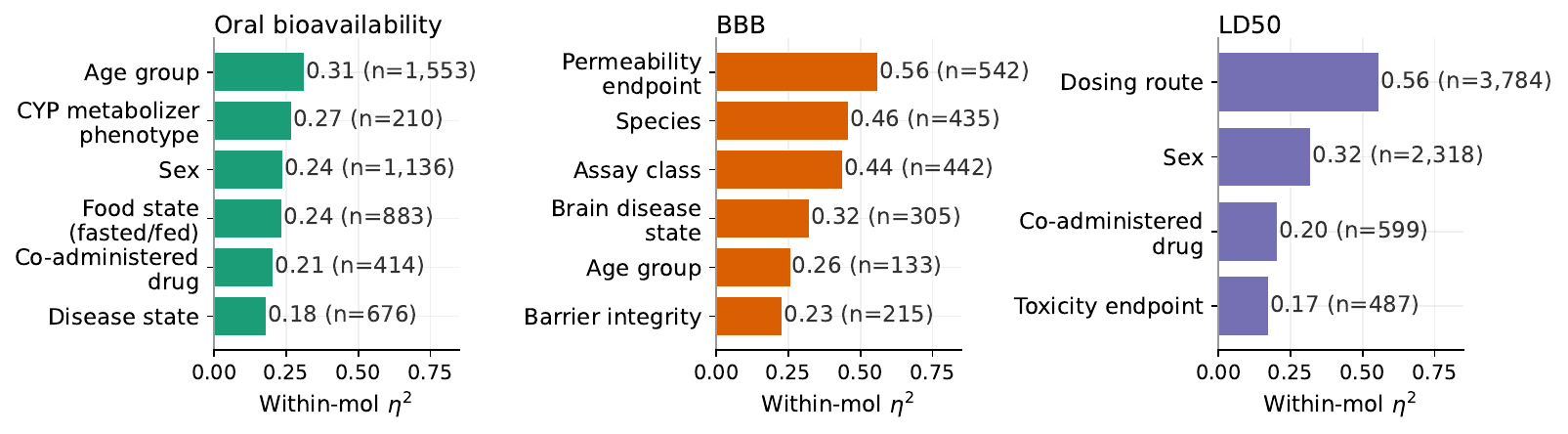}
    \caption{
Within-mol $\eta^{2}$ ANOVA effect size statistics for each task / conditioning variable: the fraction
of a molecule's label variance explained by that single covariate, averaged
across molecules with $\geq 2$ extractions covering $\geq 2$ subcategories.
For example, ``Dosing route'' splits each molecule's LD50 measurements into
$\{$oral, injection, inhalation, dermal, $\ldots\}$ and asks how much of
that molecule's lethal-dose variance falls between routes (vs.\ within a
route): $\eta^{2}=0.56$ means that over half of the variance for a single molecule's LD50 measurements in the literature is explained by route, not noise. Definitions for groups are in \cref{supp:eta2_enums}; all $\eta^2$ values significant with $p < 0.001$}
    \label{fig:within_mol_variance}
\end{figure}

\paragraph{Nuanced contexts are common and predictive, yet current datasets fail to capture them.}
\ourmethod data shows that differing experimental outcomes for the same molecule are meaningful, not just noise.
We find such cases are common: for one-third of molecules with more than five \ourmethod extractions in BBB and LD50, at least two distinct outcomes appear, each in at least 20\% of extractions. 
Existing datasets would mislabel these molecules because they only consider structure.

Such context strongly affects label distributions across many tasks and is not noise.
In \cref{fig:within_mol_variance}, we use ANOVA to quantify it's importance.
The conditioning variables are from structured context, arising from interaction between sub-agents during extraction (e.g., dosing route, food state, sex, assay class, disease state, age group) and are output alongside the target values (e.g., binary labels for Oral and BBB, and $\log_{10}(\text{mol}/\text{kg})$ for LD50)~\footnote{For this analysis, we manually filtered \ourmethod's discovered fields for those that vary within molecule. For example, we did not include ``mechanism of action'' because this nearly never varies within molecule and therefore cannot possibly explain variance in single-molecule measurements.}.
For each (task, conditioning variable) pair we restrict to molecules with $\geq 2$ extractions covering $\geq 2$ subcategories of that variable and compute the fraction of total label variance explained by each conditioning variable ($\eta^{2}$). 
All structured nuance fields proposed by \ourmethod are statistically significant and together explain a substantial proportion of variance in the observed labels.
This supports the hypothesis that such context strongly influences measurements, and that assigning each molecule a single binary label, as current datasets do, discards systematically predictable biological behavior not captured by structure alone.
\paragraph{Nuanced contexts enable better characterization of model failures and improved training.}
Much of the context we discover has known causal relationships with target properties but is entirely discarded by other resources. 
This makes it difficult to know when predictors trained on such data will succeed.
In \cref{tab:stratified}, using \ourmethod contextual data for the \texttt{BBB} and \texttt{LD50} tasks, we analyze MiniMol classifiers trained on the corresponding TDC datasets, stratified by contextual variables and evaluated on held-out \ourmethod data.
 In \texttt{BBB}, molecules cross the blood-brain barrier mainly via passive diffusion, determined by physicochemical properties, and active influx via transporters such as Pgp.
~\cref{tab:stratified} highlights the issue: TDC data poorly covers Active influx molecules, and performance on this set is at chance.
\ourmethod more uniformly covers the main diffusion mechanisms, so training on our data yields a classifier that excels at all mechanisms (+Lit).
A similar argument applies to \texttt{LD50}, where TDC poorly covers molecules that are more toxic by injection than by oral consumption.

\begin{table}[!htbp]
      \centering
      \caption{Stratified AUROC on held-out literature test sets. \textbf{(a)} BBB permeability grouped by each molecule's dominant transport mechanism from the \ourmethod{}-extracted \texttt{bbb\_transport\_label}: Passive (passive diffusion), Active influx (protein-mediated transport into the brain), and Efflux limited (active pumping out of the brain). \textbf{(b)} Rat oral LD50 grouped by the difference between oral and injection LD50 where available. In both panels, the eval target matches TDC's. In both tasks, structure-only training fails in buckets where labels are most decoupled from structure; adding the literature corpus closes most of this gap, with negligible change in structurally typical buckets.}
    \label{tab:stratified}
     \vspace{1em}
      \begin{minipage}[t]{0.48\textwidth}
          \centering
          \rowcolors{2}{white}{gray!6}
          \small
          \begin{tabular}{lrrr}
              \toprule
              \multicolumn{4}{l}{\textbf{(a) BBB by transport mechanism}} \\
              \midrule
              Mechanism         & \# Mols & TDC-only & + Lit. \\
              \midrule
              Passive diffusion & 1949 & 0.665 & \textbf{0.728} \\
              Active influx     & 1207 & 0.443 & \textbf{0.693} \\
              Efflux limited    &  897 & 0.573 & \textbf{0.727} \\
              \bottomrule
          \end{tabular}
      \end{minipage}\hfill
      \begin{minipage}[t]{0.48\textwidth}
          \centering
          \rowcolors{2}{white}{gray!6}
          \small
          \begin{tabular}{lrrr}
              \toprule
              \multicolumn{4}{l}{\textbf{(b) Rat oral LD50 by oral-vs-injection $\Delta$}} \\
              \midrule
              Oral vs Injection LD50         & \# Mols & TDC-only & + Lit. \\
              \midrule
              Injection $\approx$ Oral & 386 & 0.836 & \textbf{0.842} \\
              Injection $>$ Oral             & 383 & 0.778 & \textbf{0.781} \\
              Injection $\gg$ Oral & 384 & 0.557 & \textbf{0.706} \\
              \bottomrule
          \end{tabular}
      \end{minipage}
     \vspace{1em}
  \end{table}

\vspace{-\baselineskip}

\section{Related Work}
\label{sec:related-work}

\paragraph{Scientific research agents.} A growing class of LLM agents reads scientific literature to answer questions or execute workflows. 
PaperQA2 \citep{skarlinski_language_2024} is a representative narrative-answer system, retrieving and reranking passages to produce cited prose; Aviary \citep{narayanan_aviary_2024} extends this with a gym and training framework for FutureHouse's scientific agents, including a PaperQA-style literature environment. 
BioMNI \citep{huang2025biomni} coordinates about 150 specialist tools and 59 biomedical databases to run research workflows, using the literature mainly for tool discovery rather than as a primary data source. 
In chemistry, ChemCrow \citep{bran_chemcrow_2024} pairs an LLM with tools for synthesis planning, retrosynthesis, and literature search, while Coscientist \citep{boiko_autonomous_2023} connects an LLM to wet-lab automation hardware to design and run experiments. 
AI Scientist V1 and V2 \citep{lu_ai_2024,yamada_ai_2025} autonomously run machine-learning experiments and write papers. However, none sweep a corpus and emit \emph{structured records at literature scale}.

\vspace{-5pt}
\paragraph{LLM-based structured extraction from scientific text.} Recent work applies LLMs to extract structured records from primary literature. \citet{agrawal_large_2022} demonstrate few-shot clinical extraction; \citet{dagdelen_structured_2024} fine-tune GPT-3 to emit JSON records for materials chemistry, pursuing essentially the same goal as ours but at roughly four orders of magnitude smaller scale and with hand-crafted schemas for each task. \citet{ai_extracting_2024} fine-tune a 7B model on 100K USPTO procedures to populate the Open Reaction Database schema with high accuracy.

\vspace{-5pt}
\paragraph{Biomedical entity tagging.} Existing NER taggers like PubTator3 \citep{wei2024pubtator} and BERN2 \citep{sung_bern2_2022} support only fixed biomedical entity types via specialized models. Recent work shows that LLMs now match or surpass BERT-style taggers on biomedical NER \citep{obeidat_llms_2025}, and UniversalNER \citep{zhou_universalner_2024} introduces the teacher–student distillation paradigm we adopt at literature scale.

\vspace{-3pt}
\section{Conclusion}
\label{sec:conclusion}
\vspace{-5pt}
Manually curated biomedical databases have long underpinned therapeutic AI, but are costly to build, lag behind the literature, and discard experimental context. As models advance, these limitations is becoming a hard ceiling. We show that primary literature can instead be autonomously converted into structured datasets, cheaply and at scale. Across six therapeutic-discovery tasks, \ourmethod produces \totalextract records, yielding datasets that match or exceed the size and accuracy of the curated benchmarks they extend. The supporting passages and structured fields preserve key nuances—food state, dosing route, mechanism, assay class, disease state—that explain much of the within-molecule variance usually treated as noise. We release the datasets, code, and an API endpoint for our system.

\begin{ack}
This research was developed with funding from the Defense Advanced Research Projects Agency's (DARPA) SciFy program (Agreement No. HR00112520300) and National Science Foundation ( Grant No. IIS-2145644 and DBI-2400135). Any opinions, findings, and conclusions or recommendations expressed in this material are those of the author(s) and do not necessarily reflect the official policy or policy positions of the National Science Foundation, Department of Defense or the U.S. Government.
\end{ack}

\bibliographystyle{plainnat}
\bibliography{references}

\appendix
\crefname{appendix}{appendix}{appendices}
\Crefname{appendix}{Appendix}{Appendices}
\crefalias{section}{appendix}
\crefalias{subsection}{appendix}
\crefalias{subsubsection}{appendix}
\section{Release Plan, Limitations, and Broader Impact}
\label{sec:open_access_plan}

A core goal of this work is to make self-driving dataset construction broadly available to the research community. Two constraints make a single, monolithic release impractical:

\paragraph{Licensing.} Our $22.5$M-paper corpus is assembled under journal- and publisher-specific licensing terms that permit our internal use of the full text for AI workloads but do not extend to redistribution. We cannot release the corpus or the per-paper extracted text. We are not yet sure whether we can release the dense embeddings derived from it. Re-extracting, running OCR, and processing an equivalent corpus is possible for any institution with the appropriate licenses, but it is not a casual undertaking.

\paragraph{Engineering.} The production version of \texttt{Starling} is built on top of retrieval and indexing infrastructure designed specifically for fast querying over $250$M chunks and $4.5$B entity mentions: sharded ClickHouse bitmap indexes for entity filters, Milvus for dense vector search, custom fusion-scoring code, and ingestion pipelines for keeping the indexes consistent across re-tagging passes. Most of this complexity exists because the underlying corpus is enormous; it would be nearly all dead weight on a corpus of, say, $10$K papers, and releasing it as-is would saddle adopters with infrastructure they neither need nor can usefully replicate without the corpus it was built around.

We therefore adopt a two-pronged release strategy designed to give different audiences the right level of access while respecting both constraints.

\subsection{Hosted Access to the Full System}

The first prong preserves access to \texttt{Starling} \emph{as deployed}---running over the full $22.5$M-paper corpus, with all of its tagging, normalization, and retrieval infrastructure intact. We plan to release this as an MCP (Model Context Protocol) server that exposes \texttt{Starling}'s full end-to-end extraction pipeline to the client. Crucially, the server performs no LLM inference of its own: all model calls are issued back to the client as MCP sampling requests, which the client fulfills using its own LLM endpoint (whether a hosted API, an internal deployment, or local compute). The client pays for its own inference; we pay only for the database queries. Researchers can then use whichever LLMs they prefer, including frontier models that would be cost-prohibitive for us to serve at corpus scale.

We commit to maintaining this hosted access as long as serving costs remain tractable, which we think should be possible as the plan detailed above only requires us to serve a CPU-only database, not LLM endpoints. If aggregate query load grows beyond what we can absorb, we will introduce per-account rate limits before restricting access; the goal is to keep the system available to academic users.

\subsection{Self-Hostable Open-Source Release}

The second prong is a fully open-source version of \texttt{Starling} that any institution can stand up on a corpus they have rights to. This release ships the same agent definitions, prompts, schema-induction logic, filter-design loops, validators, and extractors used in the hosted system---the recipes are identical---so that results obtained on a self-hosted corpus are directly comparable in methodology to those obtained against ours.

We will abstract the retrieval tools to be able to run with any database backend, both the somewhat complex mixture of systems we chose for scale and simpler to stand up ones like DuckDB plus FAISS. We provide reference implementations for both small and large corpora. The API contract guarantees that \texttt{Starling} itself will run unchanged against any backend that satisfies it.

We also release the entity tagger model weights and the schemas, prompts, and configuration files for each of the six tasks reported in the main paper, so that a self-hosted deployment can reproduce our extractions on the user's own corpus, and so that practitioners can use them as templates for new extraction tasks.

\subsection{Limitations and Broader Impact}
\label{app:impacts}

\ourmethod reads text and tables but not figures, so any property reported only in a plot is currently unavailable to the agent. Accessing this data will require a multimodal extractor. Our corpus is licensed for internal use, but not for redistribution, so external groups can reproduce the methodology, but without our forthcoming API endpoint or an independent license to the same corpus, they cannot reproduce our exact results. In terms of extraction quality, they inherit the biases and errors that may be found in scientific literature, and inherit the errors that may arise when using LLMs at scale. Care was taken to minimize the errors that make it out of our pipeline, with an error rate ranging from $0.6$ to $7.78\%$ per task.

\paragraph{Data provenance.} \ourmethod{} has no opinions about data provenance or how to weight evidence or data from different papers or sources. The extraction agent takes the claims made in papers at face value. While it is therefore possible to build an understanding of the landscape for particular molecules, genes, or reactions (for example), \ourmethod{}-collected data does not meet the stringent evidentiary standard of collections like ClinVar.

\paragraph{Dual-use.} \ourmethod lowers the cost and latency of building literature-grounded biomedical datasets, which we expect to make therapeutic AI work that has historically required well-funded curation efforts accessible to academic groups and smaller labs. Several of our case studies (notably \texttt{LD50} and \texttt{Oral}) carry mild dual-use potential: the same records that accelerate therapeutic design also lower the cost of toxicity-aware harmful design. This dual-use concern is typical of AI for science systems, but is nevertheless always worth flagging. 

\paragraph{Necessary judging.} Another issue worth flagging here is the need for judging. The first pass extraction run results in many potentially lower quality extractions, and we used GPT-5.4 and distillation to effectively and cheaply combat this. The use of \ourmethod{} without effective judging could easily create a flood of low quality data given the low cost of running \ourmethod{}. In ongoing work, we hope to bake judging directly into the extraction process to more completely eliminate this potential pitfall.

\subsection{Summary}

To summarize, the following are released openly:
\begin{itemize}[leftmargin=*, itemsep=2pt, topsep=2pt]
    \item The \texttt{Starling} agent system, including all agent definitions, prompts, and orchestration code.
    \item The fine-tuned entity tagger model weights, along with the per-category prompts and few-shot examples used to train it.
    \item Reference implementations of the retrieval backend at two scales.
    \item Task specifications, induced schemas, and configuration files for all six datasets reported in the main paper.
    \item The six datasets themselves, including extracted records and the verbatim supporting passages whose excerpts fall within fair-use bounds.
    \item An MCP server providing hosted access to \texttt{Starling} running over the full PubMed corpus.
\end{itemize}
The following are \emph{not} released, and cannot be released under our current licensing terms:
\begin{itemize}[leftmargin=*, itemsep=2pt, topsep=2pt]
    \item The $22.5$M-paper corpus itself, including extracted text, markdown conversions, and dense embeddings.
    \item The production retrieval infrastructure, which is tightly coupled to internal data layouts and would not be useful to external users in isolation.
\end{itemize}
We believe this combination---hosted access to the production system plus a self-hostable open-source equivalent---makes the methodology of this paper reproducible and extensible by the research community, while honoring the licensing terms under which we obtained access to the underlying literature.

\section{Deep Research Agent Baselines}
\label{app:deep_research_agents}

We evaluate five frontier AI systems on their ability to perform open-ended biomedical data extraction from PubMed. Each system was given the same five prompts, with no constraints on methodology. The goal is to assess whether current ``deep research'' agents can autonomously produce structured, validated, PubMed-traceable scientific datasets.

\subsection{Task Descriptions}
\label{app:tasks}
Each model received the same prompt per task, mirroring the original prompts to \ourmethod{}, with minor variations for models that support CSV download requests. The five prompts are shown in Figure~\ref{fig:task-prompts}.

\begin{figure}[th]
    \begin{tcolorbox}[colback=gray!5, colframe=gray!60, title={\small Task Prompts Used for All Models}, fonttitle=\bfseries\small, boxrule=0.5pt, arc=2pt, left=4pt, right=4pt, top=2pt, bottom=2pt]
    {\scriptsize\ttfamily
    \begin{enumerate}[leftmargin=*,label=\textbf{T\arabic*.}]
        \item ``Find all molecules with reported blood-brain barrier (BBB) permeability status in PubMed.''
        \item ``Find all molecules with reported oral bioavailability in PubMed.''
        \item ``Find all known human gene-disease associations within PubMed.''
        \item ``Find all molecules with reported acute toxicity (LD50) values in PubMed.''
        \item ``Find all proteins with reported subcellular localization in PubMed.''
    \end{enumerate}
    }
    \end{tcolorbox}
    \caption{The five natural-language task prompts given to each model. Minor wording variations were used only for models that supported explicit CSV download requests.}
    \label{fig:task-prompts}
\end{figure}

\subsection{Experimental Setup}
\label{app:setup}

\paragraph{Models evaluated.}
We compare five agent configurations spanning different architectures and tool-use modalities:

\begin{enumerate}[leftmargin=*]
    \item \textbf{BioMNI-Phylo}: A sandboxed Python coding agent with iterative cell execution, access to tools such as NCBI E-utilities, ChEMBL, PubChem APIs, and NLP libraries (scispaCy). Produces execution trace notebooks alongside output CSVs.
    \item \textbf{Claude Research Mode} (Opus~4.6 Extended): A web-based deep research agent on claude.ai that produces comprehensive narrative reports. As it cannot produce data files, we pair it with a \emph{continuation agent} (Claude Code) that executes the recommended pipelines to produce CSVs. This constitutes a two-stage pipeline.
    \item \textbf{GPT-5.4 Deep Research}: A web-based deep research agent on chatgpt.com that produces comprehensive narrative reports. Like Claude Research Mode, it produces no data files; a continuation agent executes the designed pipelines. This is the only configuration that attempted genuine PubMed text mining (E-utilities + PubTator + regex extraction) for 4 of 5 tasks.
    \item \textbf{GPT-5.4 Pro Extended Thinking}: OpenAI ChatGPT's code interpreter / tool-use mode, prompted to return a downloadable CSV. This agent autonomously produced complete CSVs for 4 of 5 tasks in a single session by identifying and reformatting pre-existing databases.
    \item \textbf{Kosmos AI-Scientist} (Edison Scientific): An autonomous research agent that runs a multi-step discovery loop --- hypothesis generation, code execution against curated databases and PubMed, iterative refinement, and report writing --- and ships per-task ``master'' CSVs alongside a discovery PDF and bundled intermediate artefacts. Of the five baselines, Kosmos is the only one that performs explicit self-curation steps within its own pipeline (e.g., a redistributed-benchmark detector that flags PMC5494643, the Jain 2024 NCI/CADD QSAR file, and excludes it from the LD50 master), but it does not apply a per-row LLM judge to the published master. Like GPT-5.4 Pro, Kosmos produces final CSVs autonomously without a continuation agent.
\end{enumerate}

\paragraph{Producing usable artifacts.} GPT-5.4 Pro received a ``return a downloadable CSV'' suffix and produced data files directly; Kosmos AI-Scientist received the bare task prompt because its discovery loop ships CSVs natively. Claude Research Mode and GPT-5.4 Deep Research produced narrative reports rather than data files regardless of CSV-request phrasing, so for those two we executed the report's recommended pipeline with a local continuation agent (Claude Code, Opus~4.6, 1M-context) that downloaded the cited databases or APIs and normalized and merged them into the final CSV file.

\paragraph{Strategies employed.}  The baseline agents generally adopted one of the following strategies:

\begin{enumerate}
    \item \textbf{Download (DL)}: The agent found instances primarily by downloading pre-existing databases (e.g., B3DB for BBB, PubTator3/BioREx bulk files for GDA).
    \item \textbf{Download and Aggregate (DLA)}: The agent aggregated multiple pre-existing databases to form a larger database.
    \item \textbf{Literature extraction (LE)}: The agent performed genuine, original text mining from scientific articles.
    \item \textbf{Hybrid (H)}: The agent employed some hybrid of the first three strategies in the sample pipeline (e.g., BioMNI-Phylo on T1/T2/T4, Kosmos on T4/T5).
\end{enumerate}

We annotate the strategies taken by each agent on each task in Table~\ref{tab:strategy-attribution}. 
\label{app:strategy}

\begin{table}[h]
\centering
\caption{Strategies adopted by the deep research system across the various tasks. Notable crossovers: GPT-5.4 Deep Research's T3 (loaded the pre-computed PubTator3/BioREx bulk file rather than fresh NLP $\to$ S1), BioMNI-Phylo's T3 and T5 (dropped the API layer entirely $\to$ S3), and Kosmos AI-Scientist's T4/T5 (S2 backbone augmented with PubMed/PMC-OA full-text mining $\to$ H).}
\vspace{1em}
\label{tab:strategy-attribution}
    \rowcolors{2}{white}{gray!6}
\small
\setlength{\tabcolsep}{6pt}
\renewcommand{\arraystretch}{1.1}
\begin{tabular}{@{}lccccc@{}}
\toprule
\textbf{Agent} & \textbf{T1 BBB} & \textbf{T2 Oral} & \textbf{T3 GDA} & \textbf{T4 LD50} & \textbf{T5 PSL} \\
\midrule
BioMNI-Phylo      & H  & H  & LE & H  & LE \\
Claude Research   & DLA & DLA & DLA & DLA & DLA \\
GPT-5.4 Deep Res. & LE & LE & DL & LE & LE \\
GPT-5.4 Pro       & DL & DL & DL & DL & DL \\
Kosmos AI-Sci.    & DLA & DLA & DLA & H  & H  \\
\bottomrule
\end{tabular}
\end{table}

\subsection{Results}
\label{app:results}

Table~\ref{tab:main-results} presents aggregate dataset statistics for the collections gathered by the baseline agents across all 25 (model $\times$ task) evaluations. We organize by task and report output scale, identifier quality, PubMed traceability, and estimated data validity for each model. We note here that, because deep research systems were generally not designed for this task, the datasets collected are not without some caveats that we explore in \cref{app:main-results-notes}.

\paragraph{Validation methodology.} \label{app:validation} We audited each of the 25 (5 models $\times$ 5 tasks) outputs using automated validation scripts and manual spot-checks. Validation dimensions: row counts, identifier validity (SMILES via RDKit, UniProt/gene symbols against canonical lists), and PMID resolvability. Where an agent dropped recoverable per-row PMIDs during its own deduplication, we restore them on the agent's behalf using upstream dataset provenance (see Footnotes $^b$ and $^q$ of Table~\ref{tab:main-results}); this is a deliberately permissive choice that credits each baseline at the dataset level rather than penalizing it for losing provenance during post-processing.

\begin{table}[!htbp]
\centering
\caption{Comprehensive results for 5 models $\times$ 5 tasks. \textbf{Rows}: total output rows. \textbf{Unique}: deduplicated entities (molecules, genes, or proteins). \textbf{ID Fill}: fraction of rows with structural identifiers (SMILES for molecular tasks, UniProt/gene symbols for others). \textbf{ID Valid}: fraction of filled identifiers that pass validation. \textbf{PMID Fill}: fraction of rows with $\geq$1 PubMed ID (counts dataset-aggregator PMIDs at face value; $^b,^q$ are values we recovered on the agent's behalf when it dropped upstream PMIDs during dedup). \textbf{Valid}: agent-reported fraction of well-formed rows (no per-row LLM audit applied to baselines; the contrast with \ourmethod{}'s filtered output is discussed in \S\ref{app:main-results-notes}).}
\label{tab:main-results}
\vspace{1pt}
    \rowcolors{2}{white}{gray!6}
\small
\setlength{\tabcolsep}{3pt}
\renewcommand{\arraystretch}{1.15}
\newcolumntype{R}{>{\raggedleft\arraybackslash}X}
\begin{tabularx}{\textwidth}{@{}p{2.8cm}RRRRRR@{}}
\toprule
\textbf{Model} & \textbf{Rows} & \textbf{Unique Entities} & \textbf{ID Fill} & \textbf{ID Valid} & \textbf{PMID Fill} & \textbf{Valid} \\
\midrule
\multicolumn{7}{l}{\textbf{T1 -- Blood-Brain Barrier Permeability}} \\
\arrayrulecolor{black!35}\cmidrule(lr){1-7}\arrayrulecolor{black}
BioMNI-Phylo &  8,741 & 8,741 &  100\% & 99.9\% & 91.8\% &  $\sim$100\% \\
Claude Research & 20,177 & 8,716 & 100\% & 100\% & $\geq$38.7\%$^b$ & $\sim$100\% \\
GPT-5.4 Deep Res. & 11,004 & 6,463 & 54.6\% & 100\% & 100\% & $\sim$100\% \\
GPT-5.4 Pro & 7,796 & 7,796 & 100\% & 99.97\% & 0.18\% & $\sim$100\% \\
Kosmos AI-Sci. & 9,955 & 9,955 & 100\% & 100\% & $\geq$88.5\%$^q$ & $\sim$99\% \\
\midrule
\multicolumn{7}{l}{\textbf{T2 -- Oral Bioavailability}} \\
\arrayrulecolor{black!35}\cmidrule(lr){1-7}\arrayrulecolor{black}
BioMNI-Phylo & 14,089 & 12,763 & 90.6\% & 100\% & 99.0\% & $\sim$96\% \\
Claude Research & 4,525 & 2,007 & 100\% & 100\% & 100\% & $\sim$97\% \\
GPT-5.4 Deep Res. & 823 & 553 & 68.8\% & 100\% & 100\% & $\sim$99\% \\
GPT-5.4 Pro & 995 & 995 & 100\% & 100\% & 0\% & $\sim$99\% \\
Kosmos AI-Sci. & 32,609 & 22,707 & 100\% & 100\% & 93.8\% & $\sim$99\% \\
\midrule
\multicolumn{7}{l}{\textbf{T3 -- Gene-Disease Associations}} \\
\arrayrulecolor{black!35}\cmidrule(lr){1-7}\arrayrulecolor{black}
BioMNI-Phylo & 74,018 & 8,065 & 100\% & 98.1\% & 100\% & $\sim$84\% \\
Claude Research & 38.4M & 28,514 & 100\% & 100\% & 99.0\% & 0.92\% \\
GPT-5.4 Deep Res. & 1.45M & 23,899 & 100\% & 100\% & 100\% & $>$99\% \\
GPT-5.4 Pro & 10,961 & 5,097 & 100\% & 100\% & 0\% & $\sim$99\% \\
Kosmos AI-Sci. & 4.13M & 26,939 & 100\% & 100\% & 96.1\% & $>$99\% \\
\midrule
\multicolumn{7}{l}{\textbf{T4 -- Acute Toxicity (LD50)}} \\
\arrayrulecolor{black!35}\cmidrule(lr){1-7}\arrayrulecolor{black}
BioMNI-Phylo & 15,542 & 8,975 & 48.9\% & 100\% & 0.26\% & $\sim$93\% \\
Claude Research & 186,827 & 90,373 & 71.7\% & 100\% & 0\% & $\sim$99\% \\
GPT-5.4 Deep Res. & 1,132 & 871 & 50.4\% & 100\% & 100\% & $\sim$92\% \\
GPT-5.4 Pro & 8,507 & 8,507 & 0\% & --- & 0\% & $\sim$99\% \\
Kosmos AI-Sci. & 162,531 & 99,633 & 96.0\% & 100\% & 34.9\% & $\sim$99\% \\
\midrule
\multicolumn{7}{l}{\textbf{T5 -- Protein Subcellular Localization}} \\
\arrayrulecolor{black!35}\cmidrule(lr){1-7}\arrayrulecolor{black}
BioMNI-Phylo & 14,016 & 14,016 & 32.8\% & 99.1\% & 100\% & $\sim$97\% \\
Claude Research & 369,273 & 362,143 & 98.1\% & $>$99\% & 14.4\% & $\sim$99\% \\
GPT-5.4 Deep Res. & 1,226 & 1,147 & 49.4\% & $>$99\% & 100\% & 94--96\% \\
GPT-5.4 Pro & 580,017 & 78,981 & 96.4\% & 100\% & 0\% & $\sim$98\% \\
Kosmos AI-Sci. & 529,268 & 19,907 & 100\% & 100\% & 34.1\% & $\sim$99\% \\
\bottomrule
\end{tabularx}

\vspace{2pt}

{\footnotesize Markers $^{b,q}$ refer to PMID rescues we performed on the agent's behalf; see \S\ref{app:main-results-notes}.}
\end{table}

\begin{table}[tp]
\centering
\caption{Primary data sources used by each agent on each task. Rows of Table~\ref{tab:main-results} reference these sources; ``PubMed mining'' indicates original literature extraction (E-utilities $+$ PubTator $+$ regex).}
\label{tab:agent-sources}
\vspace{1em}

    \rowcolors{2}{white}{gray!6}
\scriptsize
\setlength{\tabcolsep}{3pt}
\renewcommand{\arraystretch}{1.15}
\begin{tabularx}{\textwidth}{@{}l >{\raggedright\arraybackslash}X >{\raggedright\arraybackslash}X >{\raggedright\arraybackslash}X >{\raggedright\arraybackslash}X >{\raggedright\arraybackslash}X@{}}
\toprule
\textbf{Task} & \textbf{BioMNI-Phylo} & \textbf{Claude Research} & \textbf{GPT-5.4 Deep Res.} & \textbf{GPT-5.4 Pro} & \textbf{Kosmos AI-Sci.} \\
\midrule
T1 BBB   & B3DB, BBBP, PubTator NER \citep{meng_curated_2021,martins2012bbb,wei2024pubtator}        & B3DB, PharmaBench, TDC, BBBP \citep{meng_curated_2021,niu2024pharmabench,huang2022tdc,martins2012bbb}        & PubMed mining          & B3DB \citep{meng_curated_2021} & B3DB, ChEMBL, PubChem \citep{meng_curated_2021,zdrazil2024chembl,kim_pubchem_2016} \\
T2 Oral  & ChEMBL, PubMed NLP \citep{zdrazil2024chembl}        & HobPre, Kim, TDC, QSAR-TK \citep{wei2022hobpre,kim2014critical,ma2008bioavailability,ollitrault2025qsar}           & PubMed mining          & Kim et al.\ 2014 \citep{kim2014critical} & ChEMBL, HobPre, Hou, Varma \citep{zdrazil2024chembl,wei2022hobpre,hou2007adme,varma2010physicochemical} \\
T3 GDA   & PubMed co-mention NLP      & CTD, DISEASES, ClinVar, GeneRIF \citep{davis2023ctd,pletscherfrankild2015diseases,landrum2018clinvar,ncbi_generif}              & PubTator3 (BioREx) \citep{wei2024pubtator}    & NCBI ClinVar \citep{landrum2018clinvar} & PubTator3, DisGeNET, CTD, PanelApp \citep{wei2024pubtator,pinero2020disgenet,davis2023ctd,martin2019panelapp} \\
T4 LD50  & PubChem HSDB, ChEMBL, PubMed mining \citep{kim_pubchem_2016,zdrazil2024chembl}      & NCI/CADD, ToxValDB, CATMoS, TDC \citep{lu2025toxacol,wall2025toxvaldb,mansouri2021catmos,zhu2009ld50}     & PubMed mining          & NICEATM/ICE \citep{bell2017ice} & PubChem ChemIDplus, ChEMBL, PubMed mining \citep{kim_pubchem_2016,zdrazil2024chembl} \\
T5 PSL   & PubMed, scispaCy NER, NCBI Gene \citep{neumann-etal-2019-scispacy}      & UniProt, HPA, GO, COMPARTMENTS \citep{apweiler_uniprot_2004,thul2017hpa,ashburner_gene_2000,binder2014compartments}      & PubMed mining          & COMPARTMENTS (7 sp.) \citep{binder2014compartments} & GOA, UniProt, PMC-OA mining, COMPARTMENTS, HPA, SubCellBarCode \citep{huntley2015goa,apweiler_uniprot_2004,binder2014compartments,thul2017hpa,orre2019subcellbarcode} \\
\bottomrule
\end{tabularx}
\end{table}

\subsection{Notes on Table~\ref{tab:main-results}}
\label{app:main-results-notes}

\paragraph{What rows do we keep for each agent?} None of the five baselines applies a per-row quality audit; rows are emitted as-is by the chosen pipeline (database download, multi-DB union, or one-shot text mining). \ourmethod{}, by contrast, runs the production judge from \cref{sec:extraction}; the headline numbers we report are post-filter. To make the comparison conservative \emph{against} \ourmethod{}, we credit every baseline row at face value in Table~\ref{tab:main-results}, and where an agent dropped recoverable PMIDs during its own deduplication we restore them on the agent's behalf ($^b, ^q$, see ``Rescued PMIDs'' below). Nevertheless, we \textit{do} only count rows where (1) a valid normalizable entity was recoverable, and (2) a valid label was given. For example for (1), for small molecule property tasks, we discard rows where we could not recover a valid SMILES string or name that we could normalize to a SMILES string. For example for (2), we discard rows that an agent extracted with \texttt{bbb\_status=unknown} or similar.

\paragraph{Downloading datasets.} On most (agent, task) cells the baseline downloads one or two curated source databases and reshapes them: BBB agents pull B3DB / BBBP / PharmaBench; Oral agents pull ChEMBL plus HobPre / Hou / Varma / Kim. The full per-cell source attribution is in Table~\ref{tab:agent-sources}. The substantive work of deciding which (entity, label) pairs are real and which papers warrant them was done upstream by the human curators, not the agent. Per-row data integrity is good across the cohort, but we do note that downloading a dataset is not directly comparable to collecting one \textit{de novo} from the literature.

\paragraph{Claimed per-row literature sometimes doesn't hold up.} We found that, in many baseline cases with high PMID fill, the citations were dataset-aggregator references rather than per-row primary literature (e.g., $4$ distinct PMIDs across Claude Research's $4{,}525$ T2 Oral rows).

\paragraph{Rescued PMIDs.} For two cells in Table~\ref{tab:main-results} ($^b$ Claude Research T1 BBB, $^q$ Kosmos T1 BBB), the agent dropped recoverable upstream PMIDs during its own deduplication; we restore dataset-level PMIDs on the agent's behalf via the preserved source-database tags, lifting both cells from $0\%$ to non-trivial PMID fill.

\newpage

\section{Case Study: Large-Scale Reaction Extraction from PubMed}
\label{sec:reactions}

We apply \ourmethod{} to chemical reaction mining across the full PubMed
corpus, producing structured reaction records (products, reactants,
catalysts, solvents, yields, conditions) from over $2$M papers in a single automated run. The agent produced 163 probes covering $630$ unique
entity-name strings, further expanded at query time via the entity-normalization hierarchy (Section~\ref{sec:ner}). Each prompt instructs the model to extract structured
reaction records using the 12-field schema in
Table~\ref{tab:schema_richness} under ``Ours'', designed to capture the full characterization of a chemical reaction.

\subsection{Results}
\label{sec:reactions:results}

Statistics for the final dataset collected by \ourmethod{} are detailed in Table~\ref{tab:extraction_scale}.

\begin{table}[th]
\centering
    \rowcolors{2}{white}{gray!6}
\caption{Statistics for the \ourmethod{} reactions dataset.}
\label{tab:extraction_scale}
\vspace{1em}
\small
\begin{tabular}{lr}
\toprule
\textbf{Metric} & \textbf{Judge-filtered} \\
\midrule
Total reaction records               & 916,948          \\
Unique PMIDs                         & 259,679          \\
Unique products                      & 507,957          \\
Unique reactions (reactant--product) & 806,584          \\
Novel products (not in 5-ref union)  & 401,055 (79.0\%) \\
\bottomrule
\end{tabular}
\end{table}

\paragraph{Reference datasets.}
We evaluate our extractions against five established reaction and compound
databases (ORD~\cite{kearnes_open_2021}, CHORISO~\cite{sabanzagil2023choriso},
OpenExp~\cite{liu2024reactxt}, Rhea~\cite{bansal2022rhea},
ReactionSeek~\cite{li2026reactionseek}); their domains, sizes, and the
resolved-product counts we use for downstream coverage analysis are
summarized in Table~\ref{tab:ref_datasets}.

\begin{table}[t]
\centering
    \rowcolors{2}{white}{gray!6}
\caption{Reference datasets used for evaluation. ``Raw'' is the dataset size as
originally reported; ``Resolved'' is the number of unique canonical product
SMILES after our RDKit canonicalization pipeline.}
\label{tab:ref_datasets}
\vspace{1em}
\small
\begin{tabular}{llrrp{3.2cm}}
\toprule
\textbf{Dataset} & \textbf{Domain} & \textbf{Num. Reactions} & \textbf{Resolved Products} & \textbf{Source} \\
\midrule
ORD          & Patents (74.5\%)      & 2.4M & 1,431,123   & Kearnes et al.\  \citeyear{kearnes_open_2021} \\
CHORISO      & Academic journals     & 2.2M       & 1,905,747   & Sabanza Gil et al.\  \citeyear{sabanzagil2023choriso} \\
OpenExp      & Patents (USPTO+ORD)   & 274K pairs        & 262,512     & Liu et al.\  \citeyear{liu2024reactxt} \\
Rhea         & Biochemistry (PubMed) & 18.3K       & 9,869       & Bansal et al.\  \citeyear{bansal2022rhea} \\
ReactionSeek & Org.\ Syntheses (LLM) & 1,740 TP entries  & 635         & Li et al.\  \citeyear{li2026reactionseek} \\
\midrule
\multicolumn{2}{l}{Combined (deduped)} & --- & 3,479,157 & \\
\bottomrule
\end{tabular}
\end{table}

\paragraph{Coverage.}
For reactions, we define coverage as the fraction of a reference dataset's
unique canonical product SMILES that appear in our extraction set:
$\text{Cov}(R) = |S_{\text{ours}} \cap S_R| \;/\; |S_R|$,
where $S_{\text{ours}}$ is our set of resolved product SMILES and $S_R$ is the
reference set. This measures how much of the known reaction product space our
extraction recovers, independent of precision.

The variation in coverage across datasets
reflects fundamental differences in source domain rather than extraction
quality.

\begin{itemize}[nosep,leftmargin=1.5em]
  \item \textbf{Rhea} 
    ($10.1\%$) has the
    highest coverage because it is the only dataset with systematic PubMed
    linkage ($18.7$K PMIDs). Our PubMed-sourced corpus naturally overlaps
    with Rhea's biochemical literature.
  \item \textbf{CHORISO} 
    ($4.9\%$)
    represents academic journal chemistry. Despite no direct paper
    linkage, we rediscover a substantial fraction
     of its $1.9$M products through
    canonical-SMILES matching, indicating substantial overlap between
    PubMed and the high-impact chemistry literature in CHORISO's upstream
    CJHIF set.
  \item \textbf{ORD} 
    ($1.6\%$) is
    predominantly patent-derived ($74.5\%$ USPTO). Since PubMed indexes
    journal articles rather than patents, low overlap is expected.
\end{itemize}

\paragraph{Reaction novelty and scale comparison.}

Of the $507{,}957$ unique resolved products in the
judge-filtered release, $401{,}055$ ($79.0\%$) are \emph{novel}---not
present in any of the five reference datasets ($3{,}479{,}157$ combined
unique products after deduplication). The reference-only products are
predominantly patent chemistry (ORD $74.5\%$ patents, OpenExp fully
patent-derived), a domain where PubMed has inherently minimal overlap.
We compare scale at the level of unique (reactant, product) pairs
(canonical-SMILES signatures): the judge-filtered release contains
$806{,}584$ such pairs against ${\sim}$$1.58$M for ORD, giving
$\sim$$0.51{\times}$ ORD (cf.\ Table~\ref{tab:datasets}, main paper).
By the same axis we are $\sim$$3.1{\times}$ OpenExp and $\sim$$82{\times}$
Rhea, our only reference with systematic PubMed linkage.

\begin{table}[h!]
\centering
    \rowcolors{2}{white}{gray!6}
\caption{Per-record fill rates of the structured schema columns each
reaction database ships, plus per-record provenance/procedure rows. ``---''
means the field does not exist in the dataset's released schema. ``Ours''
is the judge-filtered release ($916{,}948$ records). Reactants and
product fields are universally populated and omitted; their structural
form is documented in Section~\ref{sec:reactions:results}.}
\label{tab:schema_richness}
\vspace{1em}
\begin{threeparttable}
\small
\setlength{\tabcolsep}{4pt}
\renewcommand{\arraystretch}{1.05}
\begin{tabular}{lr|rrrrr}
\toprule
\textbf{Field} & \ourmethod{} & \textbf{ORD} & \textbf{CHORISO} & \textbf{OpenExp} & \textbf{Rhea} & \textbf{RxnSeek} \\
                & \textbf{(916K)} & \textbf{(2.38M)} & \textbf{(2.22M)} & \textbf{(274K)} & \textbf{(18.3K)} & \textbf{(3.96K)} \\
\midrule
reaction\_type           & 96\%  & ---    & ---             & ---    & ---    & ---     \\
reagent                  & 60\%  & 5\%    & 90\%            & ---    & ---    & ---     \\
catalyst                 & 23\%  & 13\%   & 5\%             & 11\%   & ---    & ---     \\
solvent                  & 61\%  & 63\%   & 83\%            & 92\%   & ---    & 89\%    \\
yield\_percent           & 56\%  & 44\%   & 100\%\tnote{a}  & ---    & ---    & 92\%    \\
yield\_type              & 64\%  & ---    & ---             & ---    & ---    & ---     \\
outcome\_status          & 100\% & ---    & ---             & ---    & ---    & ---     \\
temperature              & 52\%  & 21\%   & ---             & 78\%   & ---    & 91\%    \\
reaction\_time           & 52\%  & 42\%   & ---             & 90\%   & ---    & 91\%    \\
setup\_and\_notes        & 86\%  & ---    & ---             & ---    & ---    & ---     \\
\midrule
support\_text (paragraph-cited)   & 100\% & ---  & --- & 100\% & ---  & ---    \\
procedure\_text (any granularity) & 100\% & 75\% & --- & 100\% & ---  & 100\%  \\
\midrule
DOI/PMID linkage   & 100\% & 98\%   & ---  & ---    & 95\%   & 100\%   \\
\bottomrule
\end{tabular}
\begin{tablenotes}[flushleft]
\scriptsize
\item[a] CHORISO's \texttt{yield} field is populated on every record, but
$\sim$$44\%$ of values are $0.0$, indicating either failed reactions or
missing data. Non-zero yield fills $55.7\%$ of records.
\end{tablenotes}
\end{threeparttable}
\end{table}

\subsection{Schema Richness vs.\ Reference Reaction Databases}
\label{sec:reactions:richness}

Coverage tells us \emph{which} reactions our extraction recovers, but not \emph{what we know about each one}. Headline-scale tables equate ORD, CHORISO, and Rhea with our extraction, though they are structurally far simpler. Our 12-field schema, populated at $23$--$100\%$ per field on the judge-filtered release, captures conditions and nuances that these reference datasets sometimes or often discard.

Table~\ref{tab:schema_richness} compares the conditions and provenance
fields side by side---reactants and product, which all five reference
datasets carry on essentially every record (in varying form: SMILES, ChEBI
IDs, or IUPAC names per the breakdown in
Section~\ref{sec:reactions:results}), are omitted to focus on the
schema-breadth gap. Fields marked ``---'' are schema fields that \ourmethod{} chose but that are absent from the baseline dataset.

\section{Conditioning-variable Definitions for Fig.~\ref{fig:within_mol_variance}}
\label{supp:eta2_enums}

Each plain-language label in Fig.~\ref{fig:within_mol_variance} corresponds to a
single field in the LLM-curated extraction schema. Below we list the field
name (in parentheses) and the full set of subcategory values used to compute
within-molecule $\eta^{2}$. Records with null values for a given field are
excluded from that row's computation.

\paragraph{Oral bioavailability.}
\textbf{Age group} (\texttt{developmental\_stage}) — fetal, neonatal,
pediatric, adult. \textbf{CYP metabolizer phenotype}
(\texttt{metabolizer\_capacity}) — low (PM or CYP inhibitor), medium (EM),
high (UM or CYP inducer). \textbf{Sex} — male, female, mixed. \textbf{Food
state} (\texttt{food\_effect}) — fasted, fed-standard, fed-high-fat,
fed-low-fat, other. \textbf{Co-administered drug}
(\texttt{drug\_interaction}) — boolean: PK-affecting co-drug present (true)
or absent (false). \textbf{Disease state} — none, hepatic impairment,
renal impairment, diabetes, cancer, GI disease, other.

\paragraph{BBB.}
\textbf{Permeability endpoint} (\texttt{metric\_type}) — absolute CNS
concentration ($\mu$g/g brain), brain:blood ratio ($K_{p}$), CSF:blood
ratio, transfer permeability (PS, $P_{\text{app}}$), dose fraction
(\%ID/g), imaging radiotracer (PET / SPECT), other.
\textbf{Species} — mouse, rat, human.
\textbf{Assay class} (\texttt{assay\_type}) — ex-vivo tissue (post-mortem
brain homogenate), fluid sampling (microdialysis or CSF tap), in-vivo
imaging, BBB permeability assay (\textit{in situ} perfusion, intracarotid
upstream injection), behavioral / functional, \textit{in silico}, other.
\textbf{Brain disease state} (\texttt{bbb\_disease}) — none, tumor,
inflammation, ischemia (stroke), trauma (TBI), metabolic,
neurodegenerative, other.
\textbf{Age group} (\texttt{developmental\_stage}) — fetal, neonatal,
pediatric, adult.
\textbf{Barrier integrity} (\texttt{bbb\_integrity}) — intact, disrupted,
severely disrupted.

\paragraph{LD50.}
\textbf{Dosing route} (\texttt{route}) — oral, injection-systemic (IV / IP
/ SC), injection-CNS (intracerebral or intrathecal), intramuscular, topical
/ dermal, inhalation, other. \textbf{Sex} — male, female, mixed.
\textbf{Co-administered drug} (\texttt{co\_treatment}) — boolean: another
compound co-administered (true) or not (false). \textbf{Toxicity endpoint}
(\texttt{assay\_type}) — LD50 (median lethal dose), LC50 (median lethal
concentration; inhalation), fatal-dose (any reported lethal dose).

\section{Selected ``Clear'' Errors from Human Evaluation of \texttt{BBB\_Martins}}

Here, we provide selected ``obviously wrong'' mislabelings that we identified in \texttt{BBB\_Martins} during our human evaluation of \texttt{BBB\_Martins} $\leftrightarrow$ \ourmethod{} disagreement.

\begin{table*}[h]
  \centering
  \caption{Existing manually curated datasets have errors that are not just in the tail. This table presents five drugs that have well known \textit{canonical} labels for blood-brain barrier penetration but are mislabeled in the \texttt{BBB\_Martins} dataset. Included is the literature consensus label as well as the fraction of extractions bearing that label.}
  \label{tab:greatest-hits-bbb}
  \footnotesize
  \setlength{\tabcolsep}{4pt}
  \rowcolors{2}{white}{gray!6}
  \begin{tabularx}{\textwidth}{l c c c X}
    \toprule
    Drug & TDC & Lit.\ consensus & $n_{\text{lit}}$ & Mechanism / why TDC is wrong \\
    \midrule
    \textbf{Thiopental}    &
      {\color{tdcwrong} BBB$-$} & {\color{litright} BBB$+$, rate 1.00} & 81 &
      Canonical fast-acting IV anaesthetic with central GABA$_A$ action.
      \pmids{11270237, 23973449, 30147453} \\
    \textbf{Levodopa}      &
      {\color{tdcwrong} BBB$-$} & {\color{litright} BBB$+$, rate 0.91} & 2{,}503 &
      Active LAT1 transport; first-line Parkinson's drug.
      \pmids{14683550, 17181137, 22434495} \\
    \textbf{Propranolol}   &
      {\color{tdcwrong} BBB$-$} & {\color{litright} BBB$+$, rate 0.98} & 1{,}169 &
      CNS-penetrant $\beta$-blocker used for migraine and performance anxiety.
      \pmids{20170117, 21605061, 30031971} \\
    \textbf{Loperamide}    &
      {\color{tdcwrong} BBB$+$} & {\color{litright} BBB$-$, rate 0.29} & 1{,}015 &
      P-gp efflux substrate; net brain exposure is near-zero in intact BBB.
      \pmids{18203948, 20162690, 21428839} \\
    \textbf{Isoniazid}     &
      {\color{tdcwrong} BBB$-$} & {\color{litright} BBB$+$, rate 0.96} & 390 &
      First-line drug for tubercular meningitis precisely because it crosses.
      \pmids{1803699, 23529866, 353023} \\
    \bottomrule
  \end{tabularx}
\end{table*}

\section{\ourmethod Agent Details}
\label{app:agent-details}

This section covers the parts of the \ourmethod{} pipeline that are most useful for reading the rest of the paper: the role of each agent, the rubric the \texttt{Validator} and \texttt{Judge} share, the probe-construction loop, how the retrieval set is ordered for extraction, how the production judge is trained, and the cost of an extraction run. Concrete artifacts are not reproduced here; they are included in the code release alongside the agent definitions and orchestration code.

\subsection{Agent Roles}
\label{app:agent-roles}

\ourmethod{} coordinates five agents across query construction and extraction. \cref{tab:agent-roles} summarizes their inputs, outputs, and the model class each runs on.

\begin{table}[h]
    \centering
    \small
    \caption{Agent inputs, outputs, and driving model. Qwen3.5-9B is distilled from GPT-5.4.}
    \label{tab:agent-roles}
    \begin{tabularx}{\textwidth}{l X X l}
    \toprule
    Agent & Input & Output & Model \\
    \midrule
    \texttt{Proposer} & Task description; current probe set; \texttt{Validator} and \texttt{Investigator} feedback & New/revised probes (filter + semantic query); induced schema and task instantiation & GPT-5.4 \\
    \texttt{Validator} & Sampled passage or candidate extraction; rubric & Per-axis pass/fail and relevance verdict & \texttt{gpt-oss-120b} \\
    \texttt{Investigator} & \texttt{Validator}-rejected passages; current filter & Natural-language refinement suggestions & \texttt{gpt-oss-120b} \\
    \texttt{Extractor} & Retrieved window; schema; task instantiation & Structured record(s) with support passage & \texttt{Gemma 4 31B} \\
    \texttt{Judge} & Extraction; support text; rubric & Per-axis pass/fail & \texttt{Qwen3.5-9B} \\
    \bottomrule
    \end{tabularx}
\end{table}

\subsection{Judge Rubric}
\label{app:judge-rubric}

Every extraction is graded along five binary axes, only being admitted into the dataset if it fails on none of none of them. The same rubric is shared across all six tasks.

\paragraph{Support Fidelity.} \emph{Is the \texttt{support\_text} a faithful rendering of the source evidence from the paper?} Pass: the support text is well grounded in the paper, drawing from the cited paragraph or from nearby paragraphs, tables, or figures; light rewriting, paraphrasing, and reasonable conclusions from stated facts are fine as long as meaning and factual details remain faithful. Fail: the support text cannot be supported by the paper, materially distorts it, or fabricates details.

\paragraph{Task Relevance.} \emph{Does this extraction fall within the scope defined by the task?} Pass: the extraction captures exactly the type of data the task asks for. Fail: the extraction is clearly out of scope (for example, a general physiology statement or a different property entirely).

\paragraph{Molecule Attribution.} \emph{Is the named entity actually the subject of the claim in this support text?} Pass: the entity is the subject of the extracted claim and is either explicitly named in the support text, unambiguously established by immediate surrounding context, or referred to by a recognized synonym (e.g., noradrenaline for norepinephrine, acetylsalicylic acid for aspirin). Fail: the entity is not the subject of the claim, or is absent from both the support text and its immediate context.

\paragraph{Label Correctness.} \emph{Is the primary label (the key outcome of the extraction) correct given the support text and its context?} Pass: the primary outcome accurately reflects what the text states; the focus is on the headline label, not on auxiliary or descriptive fields. Fail: the primary label contradicts the text.

\paragraph{Accuracy.} \emph{Is the core factual claim of the extraction faithful to the source text?} Pass: the primary claim is supported by the text and nothing contradicts the source. This dimension is about factual faithfulness, not about whether the best possible schema field was chosen for a given value. Fail: the extraction contradicts the source or fabricates information not stated or inferable from the paper.

\subsection{Probe Loop and Schema Induction}
\label{app:probe-loop}

The probe-construction loop in \cref{sec:corpus-filter} terminates when the \texttt{Validator} estimates $\geq 80\%$ precision and a $\leq 15\%$ recall gap per probe, or when a configured probe-count cap is hit. The recall gap is estimated by sampling $50$ papers that are highly ranked by at least one probe's \texttt{semantic\_query} but excluded by every probe's entity filter, and asking the \texttt{Validator} what fraction carry relevant evidence. On the five non-\texttt{Reactions} tasks the loop converges to roughly $8$ probes per task, with \texttt{Reactions} being an exception, running from $37$ hand-authored task specifications that the loop expands into $163$ probes.

Schema induction follows once the probe set is frozen. The \texttt{Proposer} samples $100$ papers from the probe union and runs the \texttt{Extractor} on each in free-form mode (no fixed schema). The free-form records and the \texttt{Validator}'s per-axis scores on them are returned to the \texttt{Proposer}, which proposes a unified schema and a concrete \emph{task instantiation}---the natural-language description of what counts as a valid record, what to ignore, and what default assumptions to use---and re-scores. In practice the schema fields stabilize after the first pass; almost all of the iteration is the \texttt{Proposer} sharpening the task instantiation so that the \texttt{Extractor} stops emitting boundary cases the \texttt{Validator} disagrees with.

\subsection{Subcorpus Ranking}
\label{app:ranking}

Once the probe set is frozen, the union of windows that match any probe forms the candidate subcorpus. The \texttt{Extractor} processes papers in priority order, ranking each paper by a weighted sum of three signals: (i) the number of probes whose entity filter the paper satisfies (probe-hit count), (ii) the mean semantic similarity between the paper's windows and the probes' semantic queries, and (iii) the maximum such similarity over the paper's windows. The hit count rewards papers jointly relevant to multiple aspects of the task; the mean and max together favor papers that are either densely on-topic or contain a single highly-relevant passage.

\subsection{Distilled Judge Training}
\label{app:judge-training}

The production judge from \cref{sec:extraction} is Qwen3.5-9B distilled from GPT-5.4 verdicts on the five-axis rubric above. Per task we collect $5{,}000$ GPT-5.4-graded extractions ($30{,}000$ across the six tasks), split $3{,}000\,/\,1{,}000\,/\,1{,}000$ into SFT, GRPO, and held-out eval. SFT trains the judge to reproduce GPT-5.4's per-axis pass/fail labels across all tasks and is refined with GRPO.

\subsection{Extraction Cost Breakdown}
\label{app:cost}

The headline figure of $\sim\$0.001$ per kept record (\cref{sec:results-size}) is dominated by \texttt{Extractor} inference. Planning, schema induction, and judge filtering combined are an order of magnitude smaller and amortized across all kept records for a task. \cref{tab:extraction-tokens} reports the per-task token volume and B200 GPU-hours for the extraction stage; in aggregate we run $42.3$B input and $11.2$B output tokens through the \texttt{Extractor}, totaling $\sim$$3{,}233$ B200 GPU-hours.

\begin{table}[h]
    \centering
    \caption{Extraction-stage inference volume per task.}
    \label{tab:extraction-tokens}
    \begin{tabular}{lrrr}
    \toprule
    Task & Input tokens & Output tokens & GPU-hours \\
    \midrule
    \texttt{Reactions} & $33.6$B & $5.80$B & $2{,}035$ \\
    \texttt{PSL}       & $3.12$B & $2.06$B & $453$ \\
    \texttt{GDA}       & $2.60$B & $2.03$B & $431$ \\
    \texttt{BBB}       & $1.94$B & $792$M  & $197$ \\
    \texttt{Oral}      & $556$M  & $252$M  & $60.7$ \\
    \texttt{LD50}      & $537$M  & $233$M  & $56.8$ \\
    \midrule
    Total              & $42.3$B & $11.2$B & $3{,}233$ \\
    \bottomrule
    \end{tabular}
\end{table}

The dollar cost of this volume varies by an order of magnitude across deployment paths. We benchmark three: a hosted-API call to the same model class through OpenRouter ($\$0.13$/M input tokens, $\$0.38$/M output tokens), a vLLM deployment on rented Vast.ai GPUs ($\$3.85$/GPU-hour), and a vLLM deployment on internal compute ($\$1.00$/GPU-hour). The corresponding total extraction costs across all six tasks are $\$9{,}749$, $\$12{,}448$, and $\$3{,}233$, respectively.

\section{Corpus Details}
\label{app:corpus-details}
The full set of entity types tagged within our corpus are: \texttt{Anatomy}, \texttt{Antibody}, \texttt{Assay/Result}, \texttt{CellLine}, \texttt{CellType}, \texttt{ClinicalTrial}, \texttt{Disease}, \texttt{Gene}, \texttt{GeneVariant}, \texttt{GOTerm}, \texttt{Organism}, \texttt{Pathway}, \texttt{Peptide}, \texttt{Phenotype}, \texttt{Protein}, \texttt{Protein/GeneFamily}, \texttt{RNA}, \texttt{SmallMolecule}, and \texttt{SmallMoleculeClass}.

\cref{sec:entity-filters} introduces the entity filters that \ourmethod's sub-agents author; we give a concrete example here. The filter specification in \cref{fig:bbb-filter-example} is one of the probes the \texttt{Proposer} constructed for the \texttt{BBB} task.

\begin{figure}[h]
\centering
\begin{lstlisting}[language=jsonschema,backgroundcolor=\color{jsonbg},frame=single,rulecolor=\color{jsonframe},framesep=4pt,xleftmargin=4pt,xrightmargin=4pt]
{
  (*\textcolor{jsonkey}{\detokenize{"entity_groups"}}*): [
    ["SmallMolecule"],
    ["blood-brain barrier", "cerebrospinal fluid"]
  ],
  (*\textcolor{jsonkey}{\detokenize{"semantic_query"}}*): "Reported blood-brain barrier (BBB) permeability or brain
   penetration status for a named compound, including statements that the
   compound crosses/penetrates or does not cross the BBB (permeable/impermeable),
   or quantitative measures such as brain-to-plasma ratio (Kp), CSF concentration,
   logBB, in situ brain perfusion, microdialysis, PAMPA-BBB/MDCK permeability used
   as evidence of BBB penetration."
}
\end{lstlisting}
\caption{One of the \texttt{FilterSpec} probes the \texttt{Proposer} constructed for the
\texttt{BBB} task. \texttt{entity\_groups} is a CNF entity constraint (outer AND, inner OR);
\texttt{semantic\_query} is the rerank target applied to the surviving windows.}
\label{fig:bbb-filter-example}
\end{figure}

\section{Additional Results Measuring Disagreement for Individual Molecules}

\begin{figure}
    \centering
    \includegraphics[width=\linewidth]{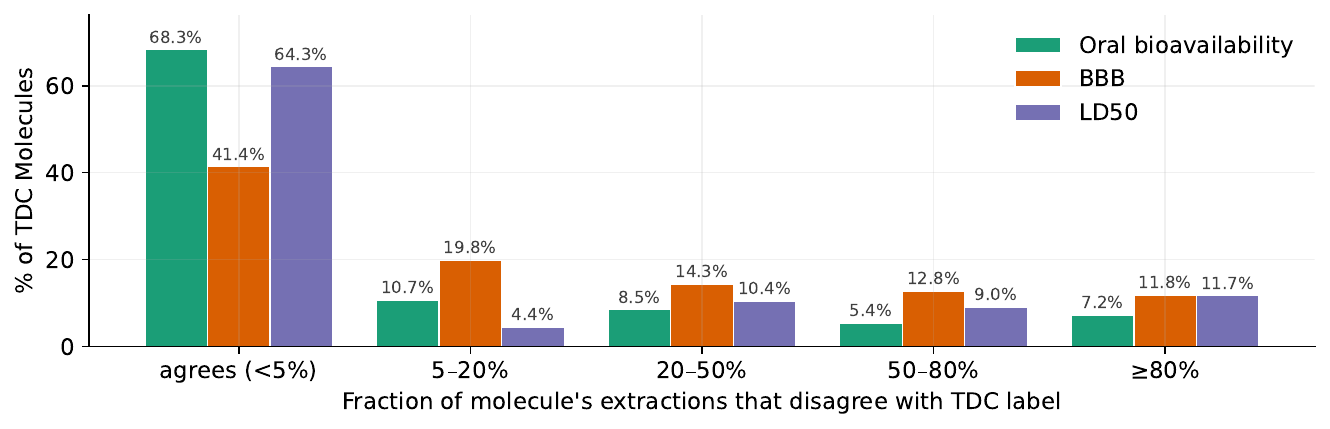}
    \caption{For each labeled molecule in three TDC datasets, what fraction of \ourmethod{}'s extractions disagree with the label? We find that between 12.6\% (Oral bioavailability) and 24.6\% (BBB) of the labels in TDC are majority disagreed with by the literature.}
    \label{fig:bucket_breakdown}
\end{figure}

In this section, we present two small case studies that measure the variability of \texttt{BBB}, \texttt{Oral} and \texttt{LD50} labels that exists for single molecules. Unlike existing resources for small molecule properties, \ourmethod{} contains many extractions for some molecules. It's reasonable to ask: how reasonable are individual labels? In \cref{fig:bucket_breakdown}, we take molecules from three TDC datasets and evaluate what fraction of \ourmethod{}'s extractions disagree with the single label for each molecule. We find that, while in two cases the majority of molecules have near-unanimous agreement in the literature, there is a significant tail of disagreement. For example, for BBB we find that 24.6\% of molecules in TDC are disagreed with by the majority of literature extractions. These disagreements comprise both errors \textit{and} genuine ambiguity due to context or legitimate biological uncertainty.  

\begin{figure}
    \centering
    \includegraphics[width=\linewidth]{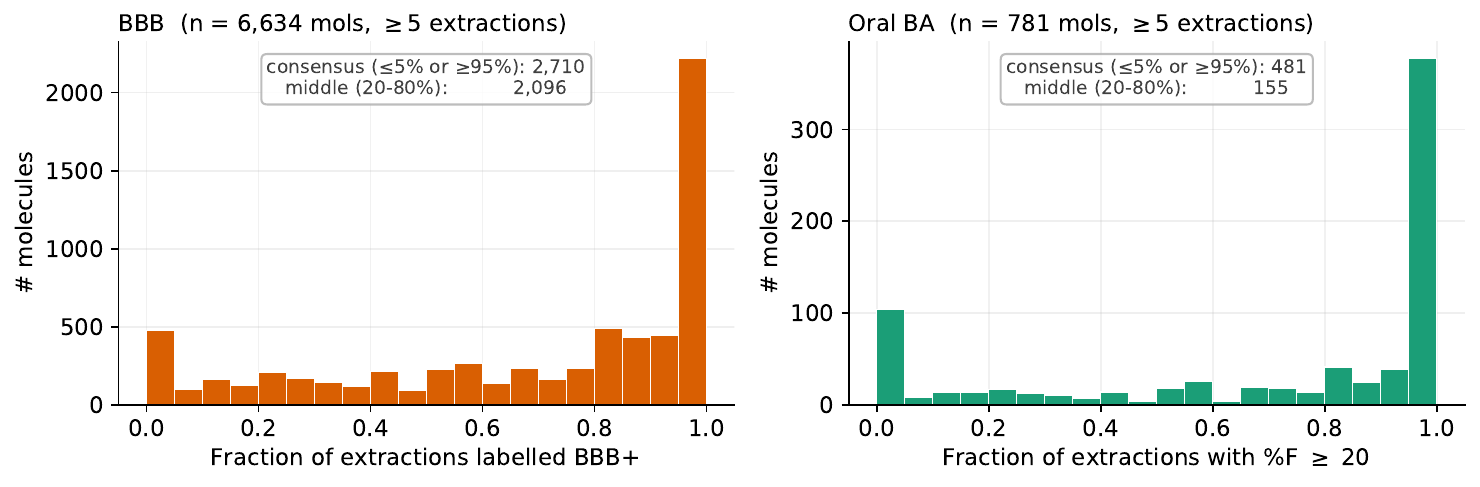}
    \caption{For each molecule with $\geq 5$ extractions, what fraction of the extractions would have the $+$ label in the corresponding TDC datasets?}
    \label{fig:positive_fraction_histogram}
\end{figure}

In \cref{fig:positive_fraction_histogram}, we show what fraction of the extractions for each \ourmethod{}-extracted molecule with $\geq 5$ extractions would have the $+$ label if they were included in the corresponding TDC datasets. Although the fully binary labels ($0.0, 1.0$) are the mode outcomes, many molecules have substantial disagreement within the literature.

\end{document}